\documentclass[12pt, a4paper]{article}
\usepackage[utf8]{inputenc}
\usepackage{enumitem}
\usepackage[breaklinks=true]{hyperref}
\usepackage[a4paper, margin=1.5cm]{geometry}
\usepackage{amsmath}
\usepackage{mathrsfs}
\usepackage{dsfont}
\usepackage{todonotes}
\usepackage{authblk}
\usepackage{acro}
\usepackage{booktabs}
\usepackage{tabularx}
\usepackage{graphicx}
\usepackage{caption}
\usepackage[labelformat=simple]{subcaption}
\usepackage{siunitx}
\usepackage{dsfont}

\newcommand{\E}{\mathds{E}}                 % Erwartungswertoperator
\newcommand{\eqdef}{\stackrel{\mathrm{def}}{=}}     % definitorisches Gleichheitszeichen
\newcommand{\NIX}[1]{}                      % Kommando zur Auskommentierung größerer Abschnitte
\NIX{
  Hinweise:
  \begin{itemize}
    \item Für Intervalle und Operatoren sollten eckige Klammern verwendet werden.
    \item Der Schätzer für die Varianz wird mit $S^{2}$ bezeichnet.
    \item Der Schätzer für die korrigierte Varianz wird mit $S^{*2}$ bezeichnet.
  \end{itemize}
} 
\DeclareAcronym{tmoi}{
  short=TMoI,
  long=target metric of interest
}

\DeclareAcronym{ci}{
  short=CI,
  long=confidence interval
}

\DeclareAcronym{ml}{
  short=ML,
  long=machine learning
}

\DeclareAcronym{mse}{
  short=MSE,
  long=mean squared error
}

\DeclareAcronym{rmse}{
  short=RMSE,
  long=root mean squared error
}

\DeclareAcronym{clt}{
  short=CLT,
  long=central limit theorem
}

\DeclareAcronym{dgp}{
  short=DGP,
  long=data generating process
}

\DeclareAcronym{sou}{
  short=SoU,
  long=sources of uncertainty
}

\DeclareAcronym{ho}{
  short=HO,
  long=hyperparameter optimization
}
\DeclareAcronym{cv}{
  short=CV,
  long=cross-validation
}
\DeclareAcronym{led}{
  short=LED,
  long=large empirical distribution
}

\newcommand{\wrt}{w.r.t.\ }

\newcommand{\pvals}{$p$-values }

\usepackage[backend=biber, style=alphabetic, sorting=nyt, backref=true]{biblatex}

\providecommand{\keywords}[1]
{
  \small	
  \textbf{\textit{Keywords---}} #1
}

\title{Beyond Point Estimates: Distributional Uncertainty in Machine Learning Performance Evaluation}
\author[1,*]{Christoph Lehmann}
\author[1]{Yahor Paromau}
\affil{Center for Scalable Data Analytics and Artificial Intelligence (ScaDS.AI) Dresden/Leipzig, Technische Universität Dresden, Germany}
\affil[*]{Corresponding author: Christoph Lehmann, christoph.lehmann@tu-dresden.de}

\date{\today}

\newcommand{\level}{u} % define symbol for quantile level

\begin{document}

\maketitle

\begin{abstract}
Machine learning models are often evaluated using point estimates of performance metrics such as accuracy, F1 score, or mean squared error. 
Such summaries fail to capture the inherent variability induced by stochastic elements of the training process, including data splitting, initialization, and hyperparameter optimization.

This work proposes a distributional perspective on model evaluation by treating performance metrics as random quantities rather than fixed values. 
Instead of focusing solely on aggregate measures, empirical distributions of performance metrics are analyzed using quantiles and corresponding confidence intervals.

The study investigates point and interval estimation of quantiles based on real-data use cases for classification and regression tasks, complemented by simulation studies for validation. 
Special emphasis is placed on small sample sizes, reflecting practical constraints in machine learning, where repeated training is computationally expensive. 
The results show that meaningful statistical inference on the underlying performance distribution is feasible even with sample sizes in the range of 10–25, while standard nonparametric confidence interval remain applicable under these conditions.

The proposed approach provides a more detailed characterization of variability and uncertainty compared to mean-based evaluation and enables a more differentiated comparison of models. 
In particular, it supports a risk-oriented interpretation of model performance, which is relevant in applications where reliability is critical.

The presented methods are easy to implement and broadly applicable, making them a practical extension to standard performance evaluation procedures in machine learning.
\end{abstract}
% target group: statistically interested people from machine learning
\section{Introduction}
\label{sec:introduction}

In \ac{ml} training, outcomes are influenced by numerous factors, including the train-test split, optimizer choice, initial weights, hyperparameter optimization etc. 
In the sense of scientific meaning, these factors are so-called confounding factors and analyzing those is considered good practice, as it enables a comprehensive understanding of the experimental conditions. 

In contrast, while confounding factors are examined in the typical \ac{ml} workflow, this is often done in an unsystematic manner, with a primary focus on achieving the best single value for a \ac{tmoi}. 
But simply selecting the best single value does not reflect the training process as an experiment, as even with a fixed choice of (hyper)parameters, results can vary due to the probabilistic nature of the modeling approach, for example due to weights initialization or data augmentation.

The objective of this work is to represent results of \ac{ml} training as a distribution of the \ac{tmoi} (such as e.g.\ accuracy or \ac{rmse}) conditional on the confounding factors. 
Drawing from a distribution of some \ac{tmoi} can be realized by using seed-controlled train runs of a \ac{ml} pipeline.
Thereby, one confounding factor is varied based on seeds, as e.g.~initial weights of a model, the train-test data split etc.
The resulting single values can be interpreted as repeated measurements from the distribution of the \ac{tmoi} for a given confounding factor, that allows to quantify its impact on the \ac{tmoi}. 
A distributional view allows for the evaluation of the training process in terms of uncertainty that can be quantified from a statistical perspective.
Generating such repeated measurements can be quite costly, as running \ac{ml} training needs time and computing resources. 
Therefore, a small to medium amount of such measurements is assumed throughout this contribution.
This amount of measurements is referred to as the sample size $n$ which will be in the order of 10-50.
Note that this sample size $n$ needs to be clearly separated from the data volume (train or test data size) within some \ac{ml} training. 
The presented methodology does not consider the data size that is used for the training itself.

\begin{figure}[!ht]
    \centering
    \begin{subfigure}[b]{0.45\textwidth}
        \includegraphics[width=0.95\textwidth]{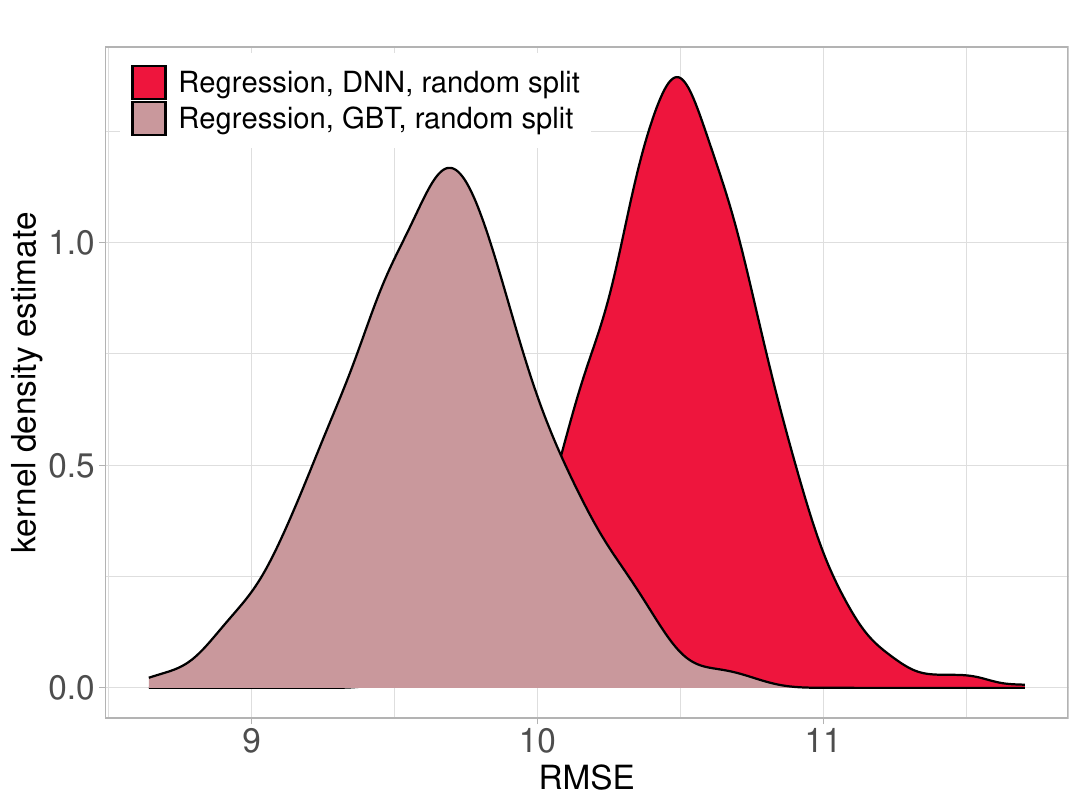}
        \caption{Empirical distributions of \ac{rmse} for regression.}
        \label{fig:intro-examples-regression}        
    \end{subfigure}
    \hfill
    \begin{subfigure}[b]{0.45\textwidth}
        \includegraphics[width=0.95\textwidth]{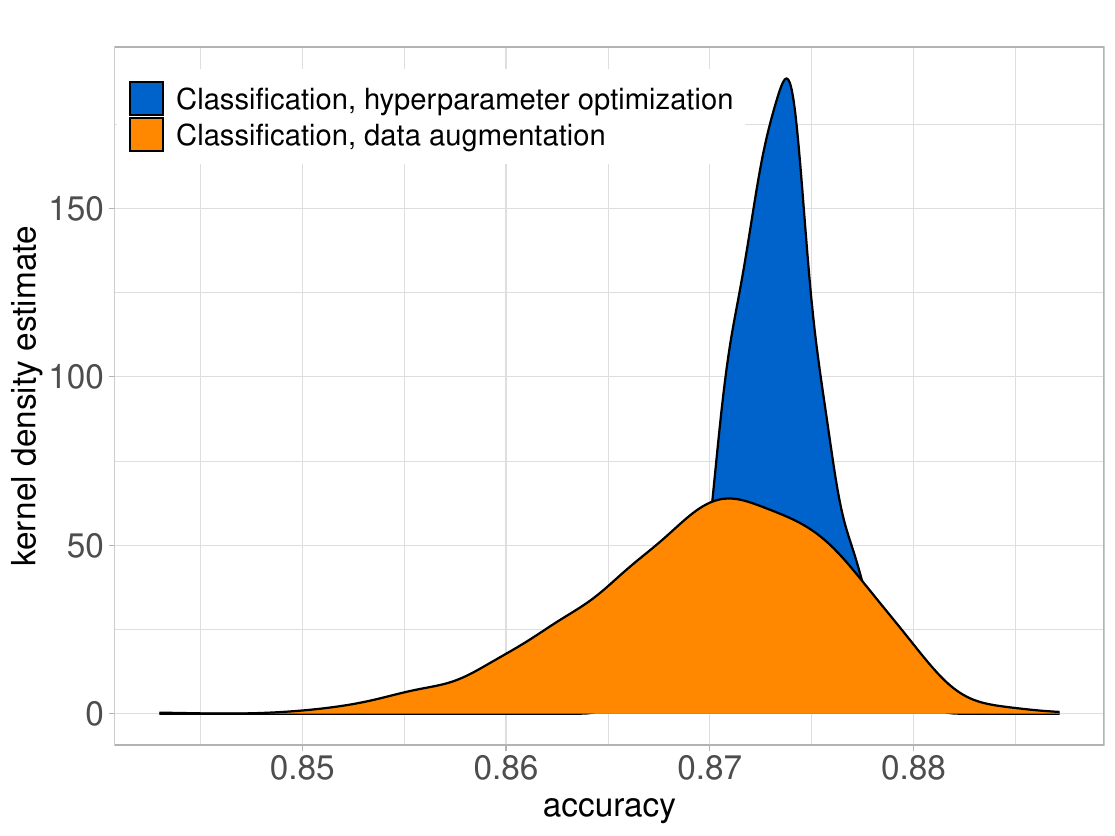}
        \caption{Empirical distributions of accuracy for classification.}
        \label{fig:intro-examples-classification}        
    \end{subfigure}
    \caption{Examples of empirical distributions of \acsp{tmoi} based on approx.~\num{1000} seed-controlled train runs.}
    \label{fig:introduction-examples-empirical-distribution}
\end{figure}

Figure~\ref{fig:introduction-examples-empirical-distribution} shows examples of empirical distributions of the \ac{rmse} for a regression problem and the accuracy rate of a classification problem.
Each of these empirical distributions is generated based on approximately \num{1000} seed-controlled train runs for different confounding factors.
Figure~\ref{fig:intro-examples-regression} shows two regression approaches for the same data (this is from the real-data use case Superconductors, see Section~\ref{sec:experiments-real-data}): deep neural network (DNN, fully connected) and gradient boosting trees (GBT). 
For both approaches, the different train runs are generated by seed-controlled train-test splits.

Figure~\ref{fig:intro-examples-classification} illustrates classifier training runs using a convolutional neural network applied to the same dataset (from the real-data use case Simpsons characters, see Section~\ref{sec:experiments-real-data}).  
One set of training runs was generated through seed-controlled hyperparameter optimization (HO), specifically using the Tree-Structured Parzen Estimator, where the seeds control initialization.  
The other set of runs was generated through seed-controlled data augmentation.  
In practice, hundreds or even thousands of repetitive measurements are rarely available.
As a consequence, the actual resulting empirical distributions consisting of observations from 10 or 25 repeated measurements typically will not look so smoothly as shown in Figure~\ref{fig:introduction-examples-empirical-distribution}.
Hence, the contribution at hand demonstrates how to work with modestly sized samples to gain insights into the unknown underlying distribution. 
Such analyses are a typical task of inferential statistics and involve considerations of uncertainty.

This contribution aims to bridge the gap between statistics and \ac{ml} by integrating statistical perspectives into \ac{ml} practice. 
When interconnecting statistics and \ac{ml}, clear definitions of terms are crucial.
Therefore, the following central terms are defined as follows. 
\begin{itemize}
    \item Theoretical Distribution: An idealized or assumed model for how probability mass is distributed.
    Such a model can be characterized by parameters or quantities as mean, variance, quantiles, etc.
    The theoretical true values of these quantities are typically unknown and need to be estimated based on data.
    \item Estimation: The process of deriving or approximating unknown quantities based on observed data.
    Estimation can result in a single value (point estimation) or a range of values (interval estimation).
    \item Uncertainty: The variability of an estimator due to sampling or algorithmic stochasticity.
    \item Confidence Interval: Estimation that provides a range of plausible values for a statistical quantity to reflect the estimation uncertainty.
\end{itemize}
Note that the above terms mainly reflect the frequentist perspective of statistics.
Bayesian approaches, in contrast, treat the quantity of interest as a random variable and rely on prior information to produce a posterior distribution. 
While Bayesian methods can provide posterior distributions for performance metrics, this contribution focuses on frequentist approaches, which are more familiar to practitioners with classical statistical training from engineering or medicine.

The core contributions of this article are as follows:
\begin{enumerate}
    \item Highlighting the stochastic nature of \acsp{tmoi} through multiple training runs, providing an engineering perspective on the number of repetitions needed for stable model evaluation.
    \item Conducting unusually large-scale repeated measurements (500–1000 runs per setting) to ensure robust empirical insights.
    \item Characterizing the empirical distribution of \acsp{tmoi} using quantiles, which allows a risk-oriented perspective on performance.
    \item Quantifying the statistical uncertainty of estimated quantiles via \acsp{ci}.    
    \item Demonstrating that the proposed approach applies to all continuous \acsp{tmoi}, including classification metrics (accuracy, precision, F1, etc.) and regression metrics (\ac{rmse}, MAE, etc.).
\end{enumerate}

Considering additional, pre-defined aspects of the evaluation distribution is of particular importance as \ac{ml} models are increasingly used in high-stakes applications such as medicine, banking, or the justice system. 
In such settings, evaluation is not only descriptive but also requirement-driven. 
Instead of relying on average performance alone, explicit acceptance criteria can be defined in advance, for example in terms of minimum performance levels or bounds on variability, which must be satisfied empirically.
This relates to a risk-oriented perspective in terms of quantiles: the 25\% quantile of accuracy indicates that the accuracy falls below this threshold in only 25\% of cases, while the 90\% quantile of RMSE shows it exceeds this threshold in just 10\% of cases.

Statistical uncertainty arises from finite samples and can be quantified using \acsp{ci}, reflecting the reliability of estimated quantiles. 
For instance, two classifiers might have similar 25\% quantiles for accuracy, but the one with the shorter \ac{ci} exhibits higher estimation precision.
To make this concrete, consider the distributions in Figure~\ref{fig:intro-examples-regression}. 
Using \acsp{ci} for the 90\% quantile of the \ac{rmse} with $n=25$ and confidence level $0.9$, the intervals are $[10.8, 11.2]$ for the DNN and $[9.8, 10.2]$ for the GBT. 
While the GBT achieves a lower overall RMSE, the comparable interval lengths indicate similar variability in the underlying distributions.

For classification (Figure~\ref{fig:intro-examples-classification}), the mean accuracies for HO and DA are $0.874$ and $0.872$, respectively, suggesting little difference at first glance. 
However, the interval lengths already indicate differences in variability, with DA showing a larger interval. 
This highlights that even for mean-based evaluation, considering \acsp{ci} instead of point estimates provides additional insight.
The picture becomes even clearer when considering the 10\% quantile with its corresponding \acsp{ci}: HO $[0.870, 0.873]$, and DA $[0.860, 0.868]$. 
The longer interval for DA signals higher variability, highlighting that HO provides more stable predictions. 
Such insights are impossible from point estimates alone and can guide model selection and indicate potential weaknesses in data or training processes.

Viewing \ac{ml} training as a controlled experiment with repeated measurements of confounding factors enables targeted investigation and evaluation of specific aspects of interest.
However, training \ac{ml} models can be costly, and creating the relevant sample sizes might be hard or even impossible. 
But even with moderate sample sizes ($n \approx 15-25$), these distributional and uncertainty-based insights become accessible. 
Smaller samples (around $n\approx10$) already allow moving beyond simple mean considerations and provide a first impression of model stability and tail behavior.

% % |experiment                                       |statistic |  n| sample_id| confidence_level|CI_type       | true_value|is_in_CI | interval_length|CI_lower |CI_upper |
% % |:------------------------------------------------|:---------|--:|---------:|----------------:|:-------------|----------:|:--------|---------------:|:--------|:--------|
% % |11_simpsons_hopt_tpe_AdamW_seeds                 |10%       | 25|       110|              0.9|np_exact_near |      0.871|TRUE     |          0.0030|0.870    |0.873    |
% % |20_simpsons_opt_split_augment_AdamW_split_sd=476 |10%       | 25|       219|              0.9|np_exact_near |      0.861|TRUE     |          0.0081|0.860    |0.868    |
% % |11_simpsons_hopt_tpe_AdamW_seeds                 |90%       | 25|       110|              0.9|np_exact_near |      0.876|TRUE     |          0.0015|0.875    |0.877    |
% % |20_simpsons_opt_split_augment_AdamW_split_sd=476 |90%       | 25|       219|              0.9|np_exact_near |      0.878|TRUE     |          0.0054|0.876    |0.881    |
% % |11_simpsons_hopt_tpe_AdamW_seeds                 |mean      | 25|       110|              0.9|t_mean        |      0.873|TRUE     |          0.0013|0.873    |0.874    |
% % |20_simpsons_opt_split_augment_AdamW_split_sd=476 |mean      | 25|       219|              0.9|t_mean        |      0.870|TRUE     |          0.0035|0.870    |0.874    |

The structure of this work is as follows. 
Section~\ref{sec:related-work} presents related work and briefly discusses two perspectives on \ac{tmoi} uncertainty.
Section~\ref{sec:statistical-background} provides the statistical background for quantile estimation, introducing two point estimators and three nonparametric approaches for estimating \acsp{ci} for quantiles of a \ac{tmoi}.
Section~\ref{sec:experiments} presents experiments on \acsp{ci} estimation for quantiles, including real-data use cases for classification and regression problems.
The section concludes with a simulation study to validate the real-data results.
Section~\ref{sec:conclusion} summarizes the main findings and suggests extensions and further work.

\section{Related Work} \label{sec:related-work}

A common approach in \ac{ml} training to improve model generalization is to apply resampling techniques, such as cross-validation. 
This process considers different train-test splits of the data and gives a more robust estimate of model performance. 
Although, this approach is primarily intended for improving generalization, it already reflects the variation regarding some \ac{tmoi}. 
But the variation thereby is mainly driven by the data and it is not possible to disentangle it from other confounding factors, as e.g.\ hyperparameter optimization, weights initialization etc.
Contributions investigating some kind of uncertainty driven by data variation, typically refer to a dedicated performance measure.
Articles targeting, e.g.\ the F1 score are \cite{takahashi2021confidenceIntF1, wang2015f1confidenceInterval}. 
Thereby, \cite{takahashi2021confidenceIntF1} follows a frequentist approach, relying on asymptotic results from the multivariate Central Limit Theorem to construct confidence intervals. 
In contrast, \cite{wang2015f1confidenceInterval} focuses on the variation of the F1 score using a Bayesian approach.
For regression problems, \cite{bayle2020crossvalidationCI} proposes confidence intervals for the mean test error of predictions. 
All these contributions, primarily analyze variation arising from data and the assumed asymptotics typically refer to the sample size of the test set.

A broader perspective, taken throughout this contribution, considers \ac{ml} training as a scientific experiment, with a focus on variation arising from different confounding factors, whereby data variation is one possible confounding factor.
Prior research has addressed this experimental perspective on \ac{ml} training, specifically investigating different sources of variation in \ac{ml}. 
According to \cite[Appendix C.1]{bouthillier2021accountingVariance}, multiple trials optimizing the learning pipeline over various sources of variation include data splits, data order (in the context of training batches), data augmentation, model initialization (initial parameter values), model stochasticity, and hyperparameter optimization. 
Many \ac{ml} approaches, such as deep learning or random forests, employ non-deterministic algorithms (e.g., SGD, Adam, AdaGrad) to solve the optimization problem, which is a further source of variation. 
A distinction between deterministic and non-deterministic approaches was pointed out e.g.\ in \cite[p.\,679]{hothorn2005design}. 
Thereby, describing the outcome of such fitting procedure (aka optimization process in \ac{ml}) as random variable aligns with the perspective in \cite{bouthillier2021accountingVariance} or \cite{philipp2018measuringStability}.
The choice of different optimization algorithms seems a process of continuous discussion, cf.\, \cite{schmidt2021-DLOptimizers}.
The relevance of these and other sources of variation depends on the methods and algorithms used.
In this context, \cite{philipp2018measuringStability} points into a similar direction, while considering methods at a smaller scale and considering regression only.
Their stability concept contains three components: algorithm, specified model, and data-generating process (DGP). 
This framework also evaluates stability without assumptions about the algorithm, specified model, or DGP, similar to \cite{bouthillier2021accountingVariance}.
In comparison to \cite{bouthillier2021accountingVariance}, there is a shift in utilizing a variation analysis not for benchmarking only, but in a more general sense to reflect about the training process for whatever \ac{ml} application. 
Additionally, the contribution at hand elaborates on the resulting distributions by analyzing them in terms of inferential statistics, more precisely by using quantiles.
In contrast, \cite{bouthillier2021accountingVariance} is employing only the mean.

Throughout this article, mainly the term source of variation for describing confounding factors, as e.g.~data sampling, augmentation, weight initialization, hyperparameter optimization etc., according to \cite{bouthillier2021accountingVariance} is used. 
An alternative term would be source of uncertainty, whereby the uncertainty notion is to be interpreted in the statistical sense.
Nevertheless, the variation term can be understood in a more general manner, without a strong relation to statistics which is maybe an advantage when addressing a more non-statistical audience.

To summarize, the current contribution is agnostic regarding the method/model and the \ac{tmoi}. 
It only assumes that repeated measurements of some arbitrary source of variation are available, typically generated through seed-controlled runs.
Furthermore, it is assumed that only a limited number of measurements is available for analysis.

Note that there is also related statistical work \wrt quantile estimation (point and interval) that is presented in the following Section~\ref{sec:statistical-background}, \nameref{sec:statistical-background}.
\section{Statistical Background}\label{sec:statistical-background}

In this section, the statistical methodology used to measure uncertainty when characterizing the distribution of the \ac{tmoi} is presented. 
Specifically, the focus is on using point estimates and \acsp{ci} of particular quantiles to quantify the distribution of the \ac{tmoi}.
Thereby, a point estimate is a single `best guess', while a \ac{ci} provides a range of values that likely contains the true quantile value.

In statistical analysis, it is common to assume an independent and identically distributed (i.i.d.) sample. 
The use of repeated measurements from seed-controlled ML runs can justify the i.i.d. assumption when varying only a single confounding factor.
In the following, the random observations $X_1, \ldots, X_n$ are assumed to form an i.i.d sample from a continuous distribution.

The statistical background section covers two main topics: point estimation of quantiles and \acsp{ci} for quantiles. 
First, the quantile definition is provided, followed by an introduction of two point estimators for quantiles. 
Specifically, the point estimators are: 1) the sample quantile, and 2) a linear interpolated estimator.
Second, there will be presented three distribution-free approaches to estimate a \ac{ci} of a quantile of a continuous distribution: 1) nonparametric exact \acsp{ci}, 2) nonparametric asymptotic \acsp{ci}, 3) semiparametric bootstrap \acsp{ci}.
For comparison reasons, the very common $t$-interval as \ac{ci} for the expectation is considered as well.

\subsection{Quantiles and Quantile Estimation}\label{sec:Quantiles-and-point-estimation}

\subsubsection*{Quantile Definition}

Given a random variable $X$ with cumulative distribution function $F(x) \eqdef P(X \leq x)$, the $\level$-th quantile~$\Tilde{x}_{\level}$ at a given probability level~$\level$ is defined as:
\begin{equation}\label{eq:definition-quantile}
  \Tilde{x}_{\level} \eqdef F^{-1}(\level) \eqdef \inf \big\{x: P(X \leq x) \geq \level\big\}, \quad \level \in\ ]0,1[.
\end{equation}
From the definition in \eqref{eq:definition-quantile} there can be derived the following probability statements:
\begin{equation}\label{eq:definition-quantile-prob-statement}
  P\left(X \leq \Tilde{x}_{\level}\right) \geq \level \quad \Longleftrightarrow \quad P\left(X > \Tilde{x}_{\level}\right) < 1-\level.
\end{equation}
Note the implications of the inequality sign in the quantile definition such that the specified probability level $u$ serves as a lower bound for statements about some \ac{tmoi} (assume that the \ac{tmoi} is described by the random variable $X$ with distribution function $F$). 

For example, in the case of RMSE as \ac{tmoi}, the 90\%-quantile corresponds to the statement: `An RMSE of $\Tilde{x}_{0.9}$ is exceeded in less than 10\% of cases', reflecting the inequality $P(X > \Tilde{x}_{0.9}) < 1-0.9$. 
With regard to an RMSE this kind of statement is fully satisfactory. 

In contrast, for an accuracy as \ac{tmoi}, the statement could be such as: `For at least 25\% of cases, the accuracy is below a level $\Tilde{x}_{0.25}$', reflecting the inequality $P(X \leq \Tilde{x}_{0.25}) \geq 0.25$. 
Note that instead of `at least', one would prefer `at most' or `less' here for considering the accuracy rate.
This can be obtained by considering the transformed random variable $-X$ instead of $X$. 
In this case, the quantiles satisfy $\tilde{x}_{\level} = -\tilde{x}_{1-\level}$. 
Consequently, the following probability statement can be derived:
\begin{equation}\label{eq:transformed-quantile-prob-statement}
P\left(X \leq \tilde{x}_{\level}\right) \geq \level 
\quad \Longleftrightarrow \quad 
P\left(-X > -\tilde{x}_{1-\level}\right) < \level.
\end{equation}
Besides a more preferable interpretation, the sign transformation will help to reduce the required sample size, in order estimate confidence intervals for quantiles that tend to the tail of a distribution.

\subsubsection*{Quantile Point Estimation}

Consider the sample mean $\bar{X} = \frac{1}{n}\sum_{i=1}^n X_i$ as a starting point, which is an estimator for the unknown expectation of a probability distribution. 
As the sample size $n$ increases, it can be shown that the sample mean has several desirable statistical properties, such as unbiasedness (on average, it equals the true expectation) and consistency (it gets closer to the true expectation as $n$ increases). 

In contrast to expectation estimation, quantile estimation is more complex. 
While the expectation gives a central tendency, quantiles are more sensitive to the distribution's shape and can vary more dramatically with a concrete sample. 
This makes quantile estimation a more challenging task.
The more extreme the quantile level of interest (e.g., 0.05, 0.01, 0.95, or 0.99), the more observations are basically necessary. 
Quantiles from the `middle' of a distribution (around 0.5) can be estimated more easily and more reliably. 
Given this effect, analyzing uncertainty in terms of quantiles requires a sufficiently large sample size.
As a consequence, in \ac{ml} applications where training takes a lot of time, an elaborated analysis of the tails of the \ac{tmoi} distribution may be not possible at all. 
Therefore, a critical reflection of available data (and sample size) to investigate the uncertainties of training results is crucial. 

If the underlying distribution is known, it is (more or less) straightforward to derive point estimators as well as \acsp{ci}. 
However, in the context of \ac{ml} metrics, it is difficult, if not impossible, to make any substantial assumptions for the distribution of the \ac{tmoi}.
As a consequence, it is assumed only that the \ac{tmoi} follows a continuous distribution with cumulative distribution function~$F$. 
The empirical cumulative distribution function $\hat{F}_n$ is an estimator of the (unknown) theoretical cumulative distribution function $F$. 

Quantiles can be estimated by so-called order statistics based on the sample quantiles.
The order statistics $X_{(i)}, i = 1, \ldots, n$ are the (random) observations $X_1, \ldots, X_n $ arranged in increasing order:
\begin{equation}\label{eq:order-statistics}
    X_{(1)} \leq X_{(2)} \leq \cdots \leq X_{(n)}.
\end{equation}
There is a relationship between sample quantiles and order statistics, as shown in \cite[p.\,305]{vanderVaart2000asymptoticStatistics}:
\begin{equation}\label{eq:quantiles-order-statistics}
  \hat{Q}(\level) \eqdef \hat{F}_n^{-1}(\level) = X_{(i)}, \quad \text{for} \ \level \in \left] \dfrac{i-1}{n}, \dfrac{i}{n} \right].
\end{equation}
Deriving quantiles based on the order statistics according to \eqref{eq:quantiles-order-statistics} is a common estimation approach.
Thereby, the expression in \eqref{eq:quantiles-order-statistics} is a step function.
To obtain a smoother behavior instead of the stepwise characteristics, the following fractional quantile estimator based on the linear interpolation of order statistics can be used (cf.\,\cite[p.\,332, eqn.\,(2.1)]{hutson2002semiparamQuantileEstBT}),
\begin{equation}\label{eq:linear-interpolation-quantile-estimator}
    \hat{Q}_L(\level) \eqdef (1 - \epsilon) X_{(\lfloor n'\level \rfloor)} + \epsilon X_{(\lfloor n'\level \rfloor + 1)}, \quad \frac{1}{n+1} < \level < \frac{n}{n+1},
\end{equation}
with $n' \eqdef n+1$, $\epsilon \eqdef n'\level - \lfloor n'\level \rfloor$ and $\lfloor\cdot\rfloor$ denoting the floor function.

For sake of simplicity and applicability we concentrate on practicability of quantile estimators and their behavior with regard to small samples, see Section~\ref{sec:experiments}.
Note that the default in Python in the numpy.quantile() function as well as in the R base quantile() function is the estimator $\hat{Q}_L$ (type=7 for both).
Furthermore, the quantile estimator $\hat{Q}$ is quite often introduced in statistical courses (and still a commonly investigated estimator).
As a consequence, the further considerations for point estimation are restricted to these two estimators mainly.

\subsubsection*{Remarks on Quantile Estimation Research}

Note that there exist many quantile estimators that are investigated under certain conditions.
Commonly, these quantile estimators show different behavior \wrt the underlying distribution and there is no single estimator that outperforms all others over all different types of distributions, cf.\,\cite{dudek2024quantileEstimators, dielman1994comparisonQuantileEstimators}.
There are different approaches to get and evaluate quantile estimators, which typically depend strongly on the distributional assumptions. 
Looking at so-called L-estimators ($\hat{Q}_L$ belongs to the class of L-estimators), that have preferable statistical properties, but so far these properties hold under the assumption of distributions belonging to a scale or location-scale family, cf.\,\cite{li2012unbiasedL-statistics}. 
Furthermore, there is still ongoing research also for the optimal order statistic as an estimator, cf.\,\cite{bieniek2023choiceQuantileEstimation} and references therein.
A more general, nonparametric framework is considered by \cite{zielinski2009optimalNonparametricQuantileEstimators}, whereby the concept of the most concentrated median-unbiased estimator is used. 
Although, the presented estimators therein are having desirable properties, the approaches are complex and not necessarily applicable for practical purposes for non-statisticians.
Furthermore, there is no clear consensus about the quality criteria of a quantile estimator (e.g. mean-unbiasedness in \cite{li2012unbiasedL-statistics}, median-unbiasedness in \cite{zielinski2009optimalNonparametricQuantileEstimators}).
Another interesting criterion especially for quantile estimation is not the (unknown) `true' quantile itself (see unbiasedness), but the probability statement that is associated with the quantile, cf.\,\cite{pajari2021probabilisticEvalOfQuantileEstimators}.

\subsection{Interval Estimation}\label{sec:non-parametric-CI}

Confidence intervals are a standard tool to quantify estimation uncertainty.
Although the focus of this work is on quantiles, the classical confidence interval for the expectation, the $t$-interval, is introduced as a reference method. 
It represents the standard approach in many \ac{ml} evaluation settings, where performance metrics are typically summarized by their mean.

Assume $X \sim N(\mu, \sigma^2)$ with unknown parameters. 
A confidence interval for $\mu = \E[X]$ is given by
\begin{equation}
  I_t = \left[\bar{X} \pm t_{n-1, 1-\alpha/2} \dfrac{S}{\sqrt{n}}\right],
\end{equation}
where $\bar{X}$ is the sample mean, $S$ the sample standard deviation, and $t_{n-1, 1-\alpha/2}$ the corresponding $t$-quantile.
Under normality, this interval achieves exact coverage $1-\alpha$.

In practice, the normality assumption may be justified for certain \acsp{tmoi}, or approximately via averaging arguments (CLT), see \cite{bouthillier2021accountingVariance}. 
Due to its robustness and reasonable performance for small sample sizes, the $t$-interval is widely used and serves as a natural baseline in the following experiments.

Analogously to the $t$-interval for the mean, confidence intervals for quantiles are used to quantify the estimation uncertainty of quantile estimates.
The focus is on approaches that do not rely on strong distributional assumptions and are therefore directly applicable in typical \ac{ml} evaluation settings.

\subsubsection{Exact Confidence Intervals}\label{sec:non-parametric-exact-CI}

The standard approach to construct an exact \ac{ci} for some quantile is based on order statistics and the binomial distribution.
For some quantile level $\level \in\ ]0,1[$, a distribution-free \ac{ci} $[X_{(k)}, X_{(l)}]$ can be constructed based on  
\begin{equation}\label{eq:CI-non-parametric-exact}
    P(X_{(k)} \leq F^{-1}(\level) \leq X_{(l)}) = r(k, l, n, \level), \quad 1 \leq k < l \leq n,
\end{equation}
with $r(k, l, n, \level) \eqdef \sum\limits_{s=k}^{l-1} \binom{n}{s} \level^s (1-\level)^{n-s}$.
Then, \eqref{eq:CI-non-parametric-exact} describes an exact \ac{ci} with confidence level $1-\alpha$ for $F^{-1}(\level)$ if $r(k, l, n, \level) = 1-\alpha$.

A practical limitation is that such an interval exists if and only if (see \cite[p.\,68, eqn.\,(1)]{zielinski2005BestExactNonparametricCI})
\begin{equation}\label{eq:CI-nonparametric-exact-existence-condition}
    \level^n + (1-\level)^n \leq \alpha.
\end{equation}
This implies a minimum sample size for every combination of quantile level~$\level$ and confidence level~$1-\alpha$. 
Some selected combinations are shown in Table~\ref{table:min-sample-size-exact-CI}.
\begin{table}[h!]
    \centering
    {\small{
    \begin{tabular}{@{}r|rrrrrr@{}}
        \toprule
        \multicolumn{1}{r}{} & \multicolumn{6}{c}{quantile level $\level$}     \\ \midrule
        confidence level $(1-\alpha)$       & 0.01 & 0.025 & 0.05 & 0.1 & 0.25 & 0.5 \\ \midrule
        0.90                                & 230  & 91    & 45   & 22  & 9    & 5   \\
        0.95                                & 299  & 119   & 59   & 29  & 11   & 6   \\
        0.99                                & 459  & 182   & 90   & 44  & 17   & 8   \\ \bottomrule
    \end{tabular}
    }}
    \caption{Required minimum sample sizes $n$ for two-sided, nonparametric exact \acsp{ci} for the $\level$-quantile at confidence level $1-\alpha$ according to \eqref{eq:CI-nonparametric-exact-existence-condition} (table is symmetric for quantile levels around 0.5).}\label{table:min-sample-size-exact-CI}
\end{table}
This has direct implications for \ac{ml} evaluation: for the small sample sizes typically available (e.g.\ $n \leq 20$), only central quantiles can be estimated reliably, whereas more extreme quantiles require substantially larger sample sizes.

Due to the discreteness of the binomial distribution, the confidence level is typically exceeded, resulting in conservative (wider) intervals.
Different construction methods have been proposed to reduce interval length while maintaining the desired coverage.
In this work, the approach proposed in \cite[pp.\,68, Section 2]{zielinski2005BestExactNonparametricCI} is applied, as it provides intervals with near-nominal coverage and reduced expected length.

\subsubsection{Asymptotic Confidence Intervals}\label{sec:non-parametric-asymptotic}

Based on asymptotic normality of the random sample quantiles, a nonparametric \ac{ci} can be constructed.
A \ac{ci} for the $\level$-th quantile $F^{-1}(\level)$ of any continuous distribution function $F$ with asymptotic confidence level $(1-\alpha) \in\, ]0,1[$ is given by, 
\begin{align}
    & P(X_{(k)} < F^{-1}(\level) \leq X_{(l)}) \approx 1 - \alpha, \quad (1-\alpha) \in \ ]0,1[\ , \quad k < l, \label{eq:CI-np-asymptotic-conflevel}\\
    & \text{with indices $k$ and $l$:} \quad  k,l = n\left(\level \pm z_{\alpha/2} \sqrt{\dfrac{\level(1-\level)}{n}} \right). \label{eq:CI-np-asymptotic}
\end{align}
Thereby, $z_{\alpha/2}$ denotes the $\alpha/2$ quantile of the standard normal distribution and $n$ is the sample size.

Since the indices $k$ and $l$ are real-valued, they do not always correspond to valid order statistics (i.e., $1 \leq k < l \leq n$) for every combination of $\level$, $n$, and $(1-\alpha)$.
Enforcing this constraint implies a minimum sample size.
Table~\ref{table:min-sample-size-np-asymptotic-CI} lists valid combinations of sample sizes, quantile levels, and confidence levels.
\begin{table}[h!]
    \centering
    \small{
    \begin{tabular}[t]{r|rrrrrrrrr}
        \toprule
        \multicolumn{1}{r}{} & \multicolumn{9}{c}{quantile level $\level$}     \\ \midrule
        confidence level $(1-\alpha)$ & 0.01 & 0.05 & 0.1 & 0.25 & 0.5 & 0.75 & 0.9 & 0.95 & 0.99\\
        \midrule
        0.90 & 446 & 87 & 42 & 16 & 7 & 9 & 25 & 52 & 268\\
        0.95 & 563 & 110 & 53 & 19 & 8 & 12 & 35 & 73 & 381\\
        0.99 & 846 & 164 & 79 & 28 & 11 & 20 & 60 & 127 & 657\\
        \bottomrule
    \end{tabular}    
    }
    \caption{Minimum values for sample size $n$ \wrt confidence level and quantile level for valid nonparametric asymptotic \acsp{ci}.}
    \label{table:min-sample-size-np-asymptotic-CI}
\end{table}
For indices that are not integer-valued, interpolation between order statistics can be applied using the linearly interpolating quantile estimator $\hat{Q}_L(\cdot)$ from \eqref{eq:linear-interpolation-quantile-estimator}.
The corresponding levels $k/n$ and $l/n$ are then used to determine the final \ac{ci} boundaries.

Asymptotic \acsp{ci} generally require larger sample sizes than exact nonparametric \acsp{ci} to achieve similar coverage.
Nevertheless, even a moderate sample size allows estimation of central quantiles such as quartiles, providing a richer perspective on the distribution than a simple mean.
This is particularly relevant in \ac{ml} evaluation, where typically only small numbers of repeated measurements are available.
However, the sample size requirements in Table~\ref{table:min-sample-size-np-asymptotic-CI} are clearly asymmetric \wrt the quantile level. 
In particular, lower quantiles require substantially larger sample sizes as there upper counterparts. 
The asymmetry stems from the constraint $1 \leq k < l \leq n$ but not from the asymptotic distribution itself, which is symmetric in $\alpha$ and $1-\alpha$. 
But the drawback of this asymmetry can be mitigated by applying the sign transformation related to the probability statement in equation~\eqref{eq:transformed-quantile-prob-statement} to the \ac{tmoi}.
For example, the sample size requirement  reduces from 42 to 25 by turning a 10\% quantile $\Tilde{x}_{0.1}$ into a 90\% quantile $-\Tilde{x}_{1-0.1}$, see Table~\ref{table:min-sample-size-np-asymptotic-CI}. 

\subsubsection{Semi-Parametric Bootstrap Confidence Intervals}\label{sec:semi-parametric-bootstrap}

Besides the above nonparametric approaches, resampling methods like bootstrapping provide a practical way to quantify uncertainty without assuming a specific distribution.  
For example, applying the nonparametric bootstrap for variance estimation and constructing \acsp{ci} is straightforward for the mean, see \cite{efron1994introductionBootstrap}.  

However, the standard nonparametric bootstrap is problematic for quantile estimation (except the median), due to an unstable bootstrap distribution \cite[Section 3.3, p.\,378]{hesterberg2015knowAboutTheBootstrap}.  
Further, for quantiles near the distribution tails (e.g., 1\%, 5\%, 95\%, 99\%), standard bootstrap intervals are constrained by the observed minimum and maximum, which is especially limiting for small sample sizes.  
A potential solution is the generalized bootstrap \cite{wang2010comparisonBTgeneralizedBTQuantileEst}, where a parametric family (e.g., generalized Laplace) is fitted to the data and bootstrap samples are drawn from this fit.  

A more flexible approach is the semi-parametric bootstrap proposed by \cite{hutson2002semiparamQuantileEstBT}, which extrapolates into the tails using a quantile function estimator.  
It only assumes a continuous distribution on the real line, in contrast to the generalized bootstrap, which relies on a parametric assumption.  
The procedure is as follows:
\begin{enumerate}
    \item Draw a random sample of size $n$ from the uniform distribution on $(0,1)$, representing quantile levels.
    \item Transform these levels into a bootstrap sample based on the original observations $X_1, \ldots, X_n$ using the semi-parametric quantile estimator $\hat{Q}_T$ (see \cite[p.\,333, eqn.\,(2.2)]{hutson2002semiparamQuantileEstBT}), 
    \begin{equation}\label{eq:bootstrap-CI-quantile-estimator}
        \hat{Q}_T(\level) \eqdef 
        \begin{cases}
          X_{(1)} + (X_{(2)} - X_{(1)}) \log(n'\level), & 0 < \level \leq \frac{1}{n+1}, \\
          \hat{Q}_L(\level),                             & \frac{1}{n+1} < \level < \frac{n}{n+1}, \\
          X_{(n)} - (X_{(n)} - X_{(n-1)}) \log(n'(1-\level)), & \frac{n}{n+1} \leq \level < 1,
        \end{cases}
    \end{equation}
    with $n' \eqdef n+1$ and $\hat{Q}_L$ the linear interpolation quantile estimator from \eqref{eq:linear-interpolation-quantile-estimator}.
    \item Compute the quantity of interest (here, a quantile at level $\level$) from this bootstrap sample.
\end{enumerate}
Repeating these steps generates the bootstrap distribution of the quantity of interest.  
The \ac{ci} is derived by selecting the corresponding percentiles from this distribution (e.g., the 2.5\% and 97.5\% percentiles for a 95\% \ac{ci}).  
The point estimate remains the sample quantile $\hat{Q}$ from \eqref{eq:quantiles-order-statistics}.  

The only distributional property required for the semiparametric bootstrap is the uniform distribution of quantiles, which is a provable result based on the probability integral transform.
This is in contrast to the generalized bootstrap method, as discussed in \cite{wang2010comparisonBTgeneralizedBTQuantileEst}, whereby the use of a generalized Laplace distribution is an assumption only. 

\section{Experiments and Validation} \label{sec:experiments}

This section describes concrete experiments for the presented point estimators and the \acsp{ci} in Section~\ref{sec:statistical-background} for various settings.
The first part analyzes concrete \ac{ml} applications, considering classification and regression problems. 
The second part validates the results from the real-data use cases by utilizing simulations for potential types of typical performance metric distributions, esp.\ distributions with bounded support on the unit interval.
These simulations should close the gap due to the more general perspective of the existing contributions dealing with \acsp{ci} of quantiles. 
The considered distributions therein rarely refer to distributions with bounded support (cp.\,\cite{hutson2002semiparamQuantileEstBT}).
Furthermore, \cite{zielinski2005BestExactNonparametricCI} does not contain any simulation studies.
For the application part and the validation part there were used mainly the same settings in terms of sample sizes and repetitions.

All experiments based on real data and simulations where run on a high performance computing cluster. 
In case of GPU utilization there were used NVIDIA A100-SXM4 Tensor Core-GPUs and NVIDIA H100-SXM5 Tensor Core-GPUs.
The simulations do not benefit from any GPU usage and were run on Intel Xeon Platinum~8470 (52~cores) @ 2.00~GHz with 512~GB~RAM.
The software stack comprises~R for the simulations and Python/Pytorch for all \ac{ml} related applications.
The relevant code of the use cases and the simulation, as well as implementations of the different types of \acsp{ci}, will be made available online.

\subsection{Real-Data Use Cases}\label{sec:experiments-real-data}

This section applies the presented point estimators and the \acsp{ci} from Section~\ref{sec:statistical-background} to the following three real-data use cases from \ac{ml} applications. 
The goal here is not to tune the models for best results in terms of the task (classification, regression), rather than analyze an experiment setting as it is.
\begin{enumerate}[label=(\alph*)]
    \item \emph{Simpsons characters}: The Simpsons use case contains an image classification task focused on recognizing 20 different characters from \emph{The Simpsons} series.
    The dataset for this task is publicly available at \href{https://www.kaggle.com/datasets/alexattia/the-simpsons-characters-dataset/versions/3}{https://www.kaggle.com/datasets/alexattia/\-the-simpsons-characters\--dataset/\-versions/3}.
    \item \emph{CIFAR10}: The CIFAR10 use case is an image classification task involving the categorization of images into one of 10 classes, such as airplanes, cars, and animals.  
    This dataset is widely used as a benchmark in \ac{ml} and is available at \url{https://www.cs.toronto.edu/~kriz/cifar.html}
    % {https://www.cs.toronto.edu/\textasciitilde kriz/cifar.html}.  
    This use case is also investigated in \cite{bouthillier2021accountingVariance}, wherein an order of 200 training runs is performed to investigate the uncertainties of different sources of variation. 
    \item \emph{Superconductors}: The superconductors use case is a regression task and targets predicting the critical temperature of a superconductor.     
    The main task here is to find the relevant features for the prediction task.
    The superconductors dataset can be found at \url{https://doi.org/10.24432/C53P47}.
    Details about the physical background can be found in \cite{Hamidieh2018superconductor}. 
\end{enumerate}
Table~\ref{tab:overview-real-data-uses-cases} gives an overview of the most important properties of the considered use cases.
\begin{table}[!ht]
    \small{
    \resizebox{\textwidth}{!}{%
    \begin{tabular}{|l|l|l|l|}
        \hline
        \textbf{Property}      & \textbf{Simpsons characters} & \textbf{CIFAR10}         & \textbf{Superconductors} \\ \hline
        \textbf{Data} &
          \begin{tabular}[c]{@{}l@{}}20 characters, \\ $\approx \num{200}$ - $\num{4000}$ images per class\end{tabular} &
          60k images, 10 classes &
          \begin{tabular}[c]{@{}l@{}}$\approx 21k$ observations, \\ 81 features\end{tabular} \\ \hline
        \textbf{Task}          & Classification               & Classification           & Regression               \\ \hline
        \textbf{\ac{tmoi}}     & Accuracy                     & Accuracy                 & RMSE                     \\ \hline
        \textbf{Method}        & CNN: simple, VGG16           & CNN: VGG11               & DNN, GBT                 \\ \hline
        \textbf{SoV} &
          \begin{tabular}[c]{@{}l@{}}Train-test split,\\ initial weights,\\ data augmentation,\\ dropout,\\ hyperparameter optimization\\ (Tree-Structured Parzen Estimator)\end{tabular} &
          \begin{tabular}[c]{@{}l@{}}Train-test split,\\ initial weights,\\ data augmentation,\\ dropout,\\ data order,\\ hyperparameter optimization\\ (Tree-Structured Parzen Estimator)\end{tabular} &
          \begin{tabular}[c]{@{}l@{}}Train-test split,\\ initial weights,\\ dropout,\\ data order,\\ row subsampling (GBT),\\ hyperparameter optimization\\ (Tree-Structured Parzen Estimator)\end{tabular} \\ \hline
        \textbf{\#SoV}         & 5                            & 6                        & 6                        \\ \hline
        \textbf{\#repetitions} & 500 (2x), 1000 (21x)         & 500 (1x), 1000 (5x)      & 500 (1x), 1000 (12x)     \\ \hline
        \textbf{\#experiments} & 23                           & 6                        & 13                       \\ \hline
    \end{tabular}%
    }
    \caption{Overview of use cases and their properties (CNN - Convolutional Neural Network, DNN - Deep Neural Network, GBT - Gradient Boosting Tree, SoV - Source of Variation)}
    \label{tab:overview-real-data-uses-cases}
    }
\end{table}

For all use cases, different sources of variation were evaluated by performing approximately \num{500} to \num{1000} seed-controlled repetitions for each source.
Note that this is a comparable high number of repetitions (compare to 200 repetitions in \cite{bouthillier2021accountingVariance}) and it required a total of approx.~\num{19000} GPU hours.

A so-called `experiment' in Table~\ref{tab:overview-real-data-uses-cases} means running a number of seed-controlled repetitions for some source of variation for fixed values of other confounding factors. 
Thereby, some sources of variation, as
e.g. hyperparameter optimization, allow for multiple settings such that for every chosen optimizer an amount of repetitions is run. 
Thus, there can be more experiments than sources of variation.
All experiments are conducted independently, i.e.\ each run starts from scratch without being influenced by the results of other runs.

\begin{figure}
    \centering
    \begin{subfigure}[b]{0.32\textwidth}
        \includegraphics[width=0.95\textwidth]{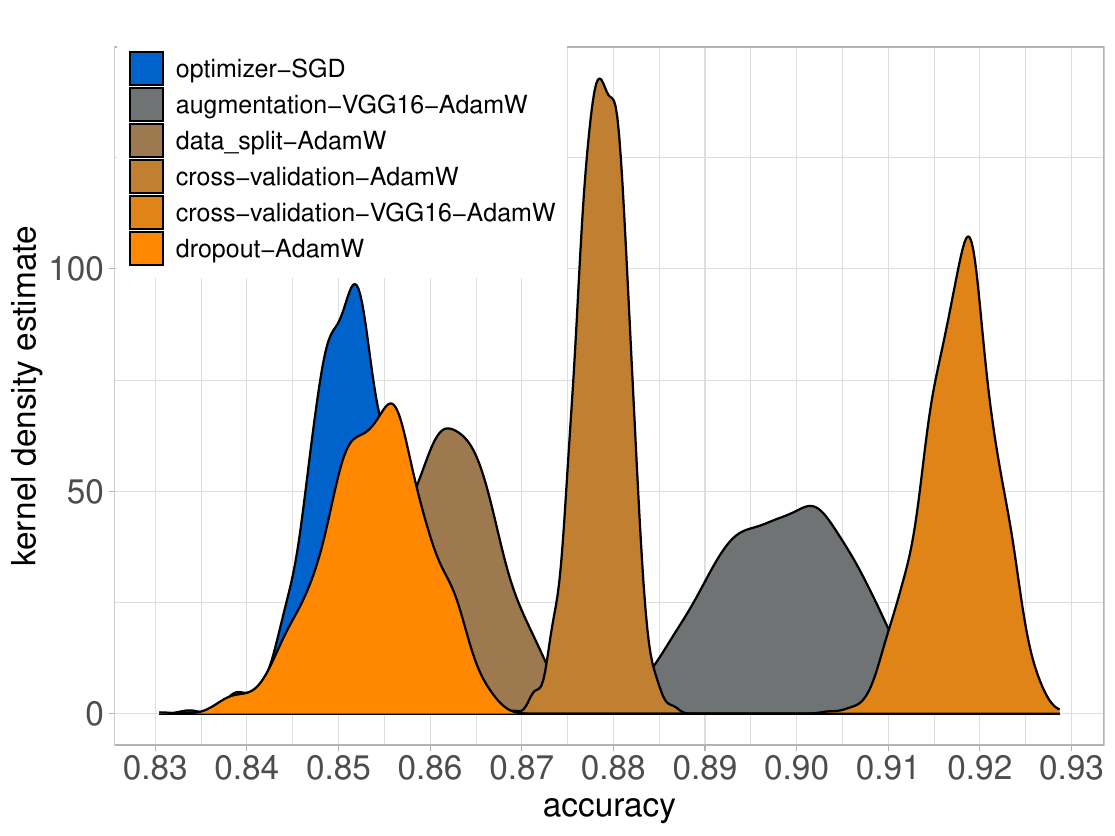}
        \caption{Simpsons characters}
        \label{fig:experiments-classification-simpsons}        
    \end{subfigure}
    \hfill
    \begin{subfigure}[b]{0.32\textwidth}
        \includegraphics[width=0.95\textwidth]{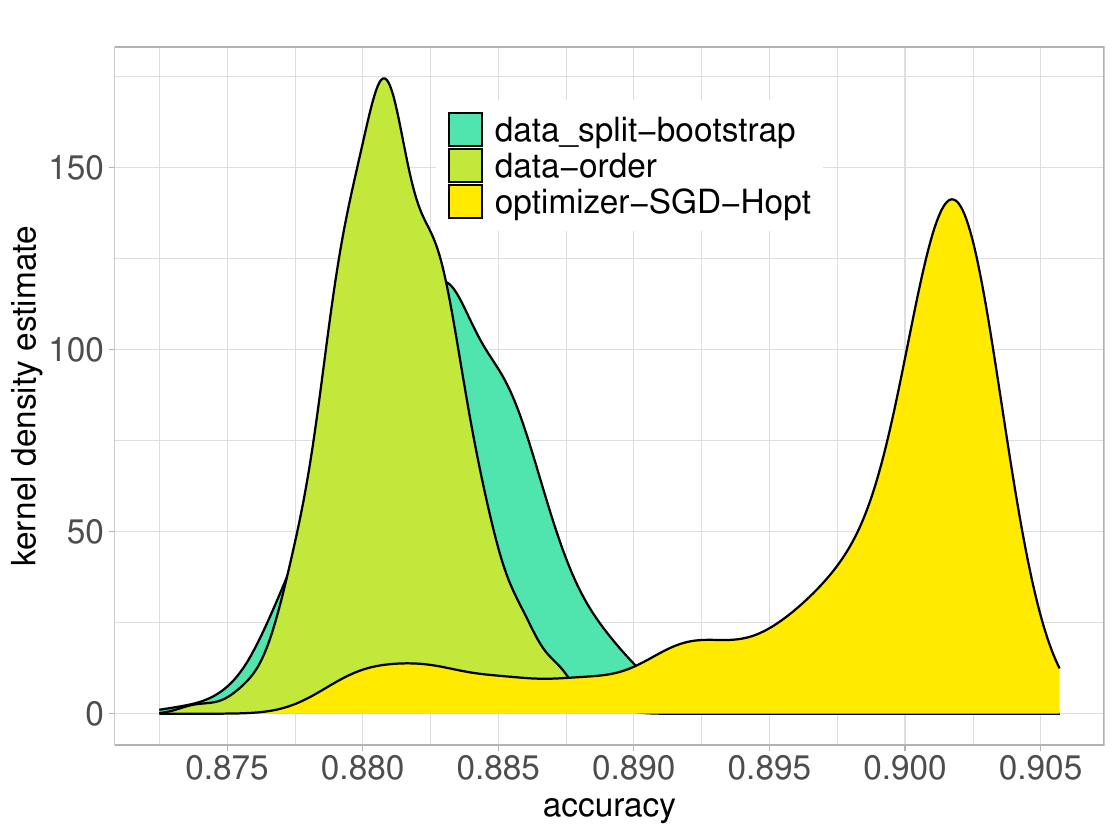}
        \caption{CIFAR10}
        \label{fig:experiments-classification-cifar10}        
    \end{subfigure}
    \hfill
    \begin{subfigure}[b]{0.32\textwidth}
        \includegraphics[width=0.95\textwidth]{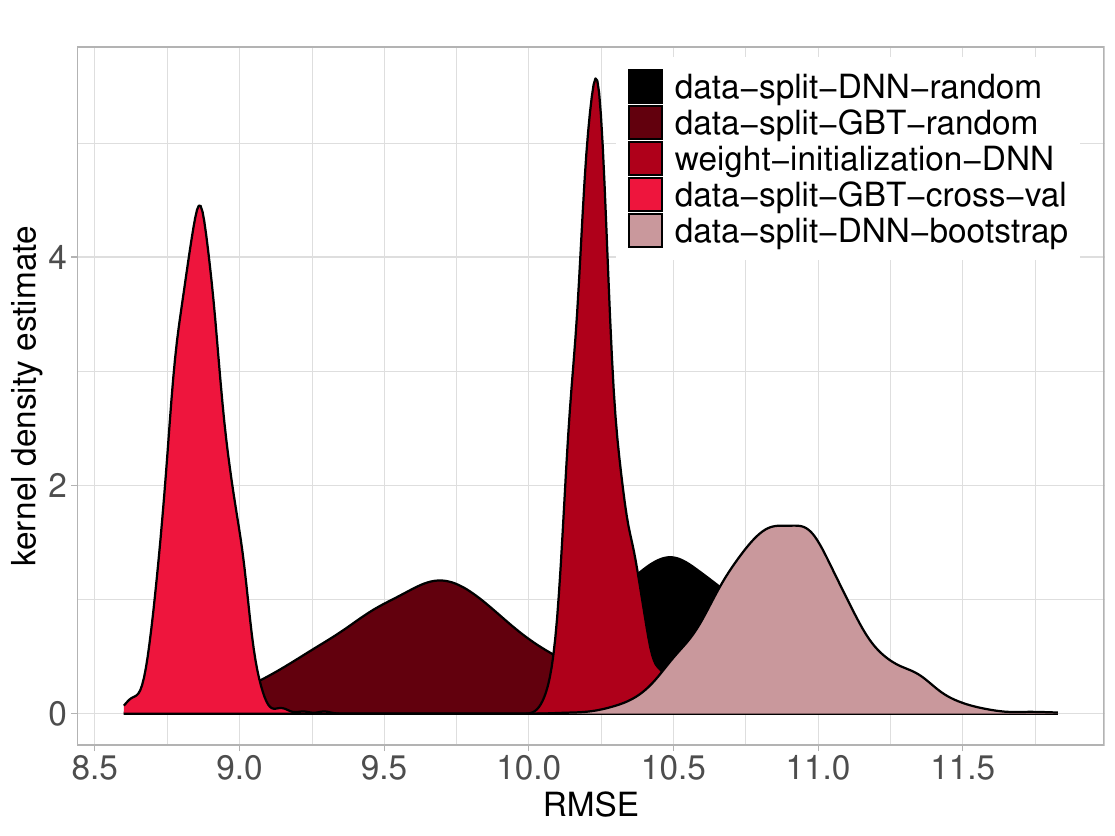}
        \caption{Superconductors}
        \label{fig:experiments-regression-superconduct}        
    \end{subfigure}
    \caption{Examples of empirical distributions of \acsp{tmoi}, each distribution based on approx.~\num{1000} seed-controlled train runs.}
    \label{fig:experiments-selected-distributions}
\end{figure}

The result of every experiment is the empirical distribution that is generated by the corresponding seed-controlled repetitions.
The kernel density estimates of some selected empirical distributions over all use cases are shown in Figure~\ref{fig:experiments-selected-distributions}.
The following evaluations assume that the true, unknown quantities (here: quantiles and mean) can be well approximated by the corresponding estimates derived from these quite extensive distributions. 
Thus, these large empirical distributions are comparable to the role of theoretical distributions in the simulations in Section~\ref{sec:experiments-simulation}. 
Consequently, the point estimate from the large empirical distribution is used as a reference for interval estimates. 
For quantile estimation, the sample quantile is used according to \eqref{eq:quantiles-order-statistics}.

Finally, an important hint on the seed control.
The sources of variation are controlled by setting the state of the corresponding random number generator (RNG) used by the functions being invoked. 
Identifying the relevant RNG can be a tricky task because different libraries utilize different RNG implementations. 
For example, Scikit-learn relies on NumPy's RNG (\verb|numpy.random.RandomState|), and as a result, most of its functions accept either an instance of \verb|numpy.random.RandomState| or an integer seed. 
On the other hand, PyTorch employs its own RNG implementation (\verb|torch.Generator|). 
Functions in PyTorch rely on the \verb|torch.Generator| instance, either accepting a specific instance or using the global (per-device) instance, as seen in operations like Dropout. 
To ensure reproducibility in PyTorch, it is also necessary to explicitly enable deterministic algorithms by using \verb|torch.use_deterministic_algorithms|. 
This is because certain operations may exhibit non-deterministic behavior due to hardware-level optimizations or asynchronous execution.
Additionally, one might control the state of Python's native RNG using \verb|random.seed|. 
In this work, a total of seven RNGs are involved and used.
Overall, careful investigation and identification of relevant seeds is a crucial part of such experiments.

\subsubsection{Quantile Point Estimation}

The behavior of the point estimators introduced in Section~\ref{sec:Quantiles-and-point-estimation} is investigated in this section \wrt two basic criteria:
\begin{enumerate}
    \item Bias: the average difference between the estimator and the true value (here: the true quantile value).
    \item MSE/RMSE: combines bias and variability (spread) of the estimates relative to the true value.
\end{enumerate} 
Based on the criteria above, an estimator with zero bias (unbiased) or less bias and a small RMSE is preferred.
Note that calculating both bias and RMSE requires knowledge of the true (unknown) quantile values. 
Here, the `true' quantile values are approximated by the sample quantile estimates derived from the extensive empirical distributions (up to \num{1000} observations).

A estimator's behavior is empirically analyzed \wrt bias and RMSE across all settings of the three use cases, using \num{2000} simulation runs. 
Within a single simulation run there are drawn random samples of size 10, 15, 25, 50 from each of the extensive empirical distribution of an experiment setting and the corresponding point estimates for the quantiles are computed.
For sake of simplicity, the experiment settings of every use case are aggregated to compute bias and RMSE of a quantile estimator.

\begin{figure}[ht!]
    \centering
    \includegraphics[width=0.95\linewidth]{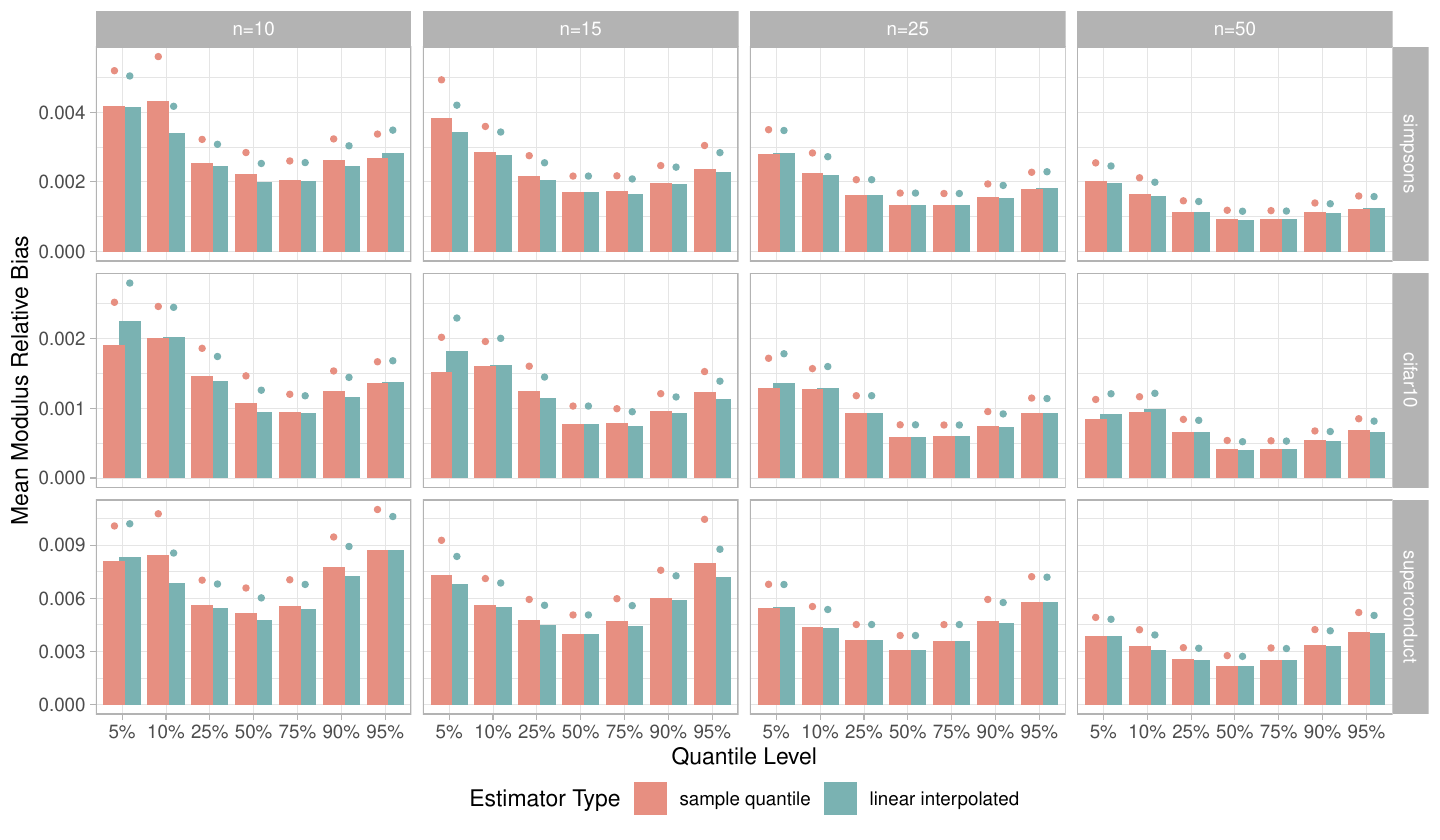}
    \caption{Mean modulus relative bias of different quantile point estimators: sample quantile ($\hat{Q}$), interpolated quantile ($\hat{Q}_L$). Points are indicating the corresponding relative RMSE. Shown results based on \num{2000} simulation runs for different quantile levels.}
    \label{fig:all-use-cases-point-estimates-bias}
\end{figure}

Figure~\ref{fig:all-use-cases-point-estimates-bias} shows the modulus bias and RMSE normalized to the true quantile values for various point estimators: sample quantile ($\hat{Q}$), and interpolated quantile ($\hat{Q}_L$). 
The modulus relative bias is represented as a bar plot, with the corresponding relative RMSE indicated by dots. 
The colors represent the different estimators. 
Figure~\ref{fig:all-use-cases-point-estimates-bias} shows the following key observations.
As the sample size $n$ increases, both (modulus) bias and RMSE decrease overall. 
Along the quantile levels, a (skewed) U-shape is observed, with lower relative bias and RMSE in the middle quantiles. 
This illustrates the earlier point that quantiles are easier to estimate in the 'middle' of a distribution than in the tails. 
Moreover, upper quantiles are slightly easier to estimate than lower ones, with relative bias and RMSE being higher for lower quantiles.
As the overall relative bias is quite small (with a maximum bias of approximately 1\%), there is little substantial or practical difference between the various estimators.
Furthermore, the relative RMSE is in a comparable order for all estimators as well.

\subsubsection{Quantile Interval Estimation}\label{sec:experiments-real-data-interval-estimation}

While point estimates provide a first indication of the performance of a model, they do not capture the inherent variability caused by the stochastic nature of machine learning training procedures. 
Therefore, it is essential to quantify the uncertainty of estimated quantiles rather than relying on point estimates alone.

This section investigates the statistical properties of \acsp{ci} for quantiles in the context of empirical performance distributions obtained from the real-data use cases. 
The goal is to assess to what extent reliable uncertainty quantification is possible based on small sample sizes, as typically encountered in \ac{ml} experiments.

The evaluation of interval estimates is based on two key criteria:
\begin{enumerate}
    \item Empirical confidence level: the proportion of intervals that contain the true quantile value.
    \item Interval length: the width of the interval as a measure of estimation precision.
\end{enumerate}
A desirable interval estimator achieves an empirical confidence level close to the nominal level while maintaining a short interval length. 
These objectives are conflicting: higher confidence levels generally result in longer intervals, whereas shorter intervals tend to reduce coverage. 
Hence, a trade-off between reliability and precision must be considered.

The empirical evaluation is conducted analogously to the point estimation study. 
For each repetition, random samples of size $n = 10, 15, 25, 50$ are drawn from the empirical performance distributions and the corresponding \acsp{ci} are computed. 
The considered methods include the exact and asymptotic nonparametric \acsp{ci}, as well as semiparametric bootstrap intervals.
Since the true quantile values are unknown, again they are approximated by the empirical quantiles derived from the extensive distributions (up to \num{1000} observations). 
This allows assessing the coverage property of the different approaches under realistic conditions.

When applying bootstrapping, the number of bootstrap samples is particularly relevant for stable estimation of \acsp{ci}. 
For variance estimation of the mean, a size of $500$ to \num{2000} is generally sufficient according to \cite[Chapter 14]{efron1994introductionBootstrap}, whereas \cite{hesterberg2015knowAboutTheBootstrap} recommends up to 10k replicates. 
Previous studies on bootstrap-based quantile estimation \cite{wang2010comparisonBTgeneralizedBTQuantileEst, hutson2002semiparamQuantileEstBT, wei2015quantileEstimatVSCIcoverage, hu2008bootstrapQuantileEstImportanceSampling, nagaraja2020distributionFreeApprox} commonly used \num{1000}–\num{2000} bootstrap samples. 
In \cite[p.\,336]{hutson2002semiparamQuantileEstBT}, \num{1000} bootstrap samples were generated for sample sizes $n=10$ and $n=25$ at a confidence level of $1-\alpha=0.95$. 
The present study extends this analysis by considering sample sizes $n=10, 15, 25, 50$ and confidence levels $1-\alpha = 0.9, 0.95$. 
To ensure comparability, simulations were performed with $\num{1000}$, $\num{2000}$, and $\num{5000}$ runs, and bootstrap intervals were computed using $\num{1000}$, $\num{2000}$, $\num{5000}$, and $\num{10000}$ bootstrap samples. 
But these variations had negligible impact on the results.
As a consequence, the presented results below are based on \num{2000} simulation runs and \num{2000} bootstrap samples, which provide stable results.

\begin{figure}[ht!]
    \centering
    \includegraphics[width=0.98\textwidth]{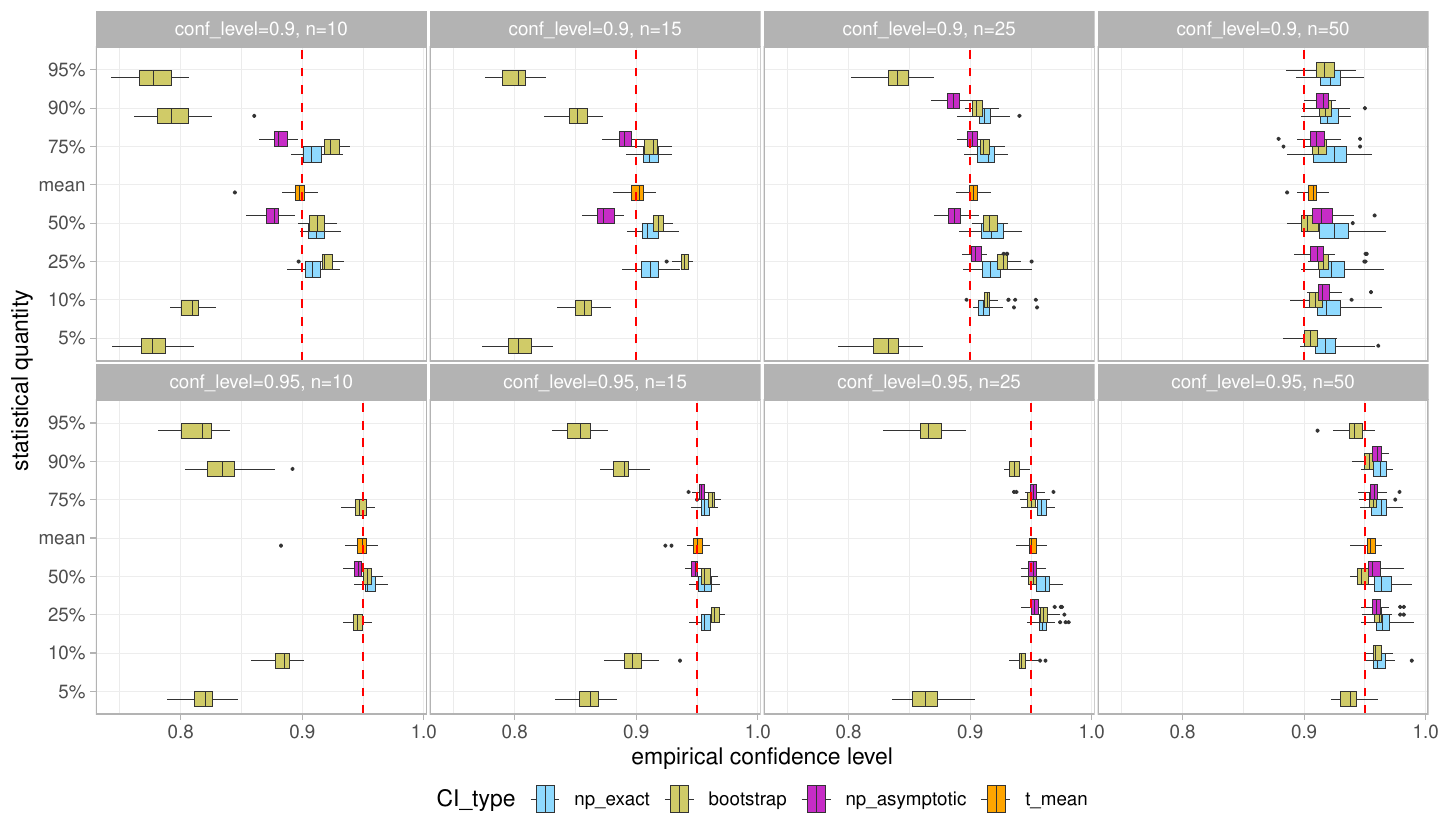}
    \caption{Empirical confidence level for different types of estimated \acsp{ci} for quantile levels $5\%, 10\%, 25\%, 50\%, 75\%, 90\%, 95\%$ and the mean for confidence levels $1-\alpha = 0.90, 0.95$ and sample sizes $n=10, 15, 25, 50$. Shown results based on \num{2000} samples of size $n$ and \num{2000} bootstrap samples. A total of 42 experiments is considered.}
    \label{fig:real-data-CI-emp-coverage}
\end{figure}

\begin{figure}[ht!]
    \centering
    \includegraphics[width=0.98\textwidth]{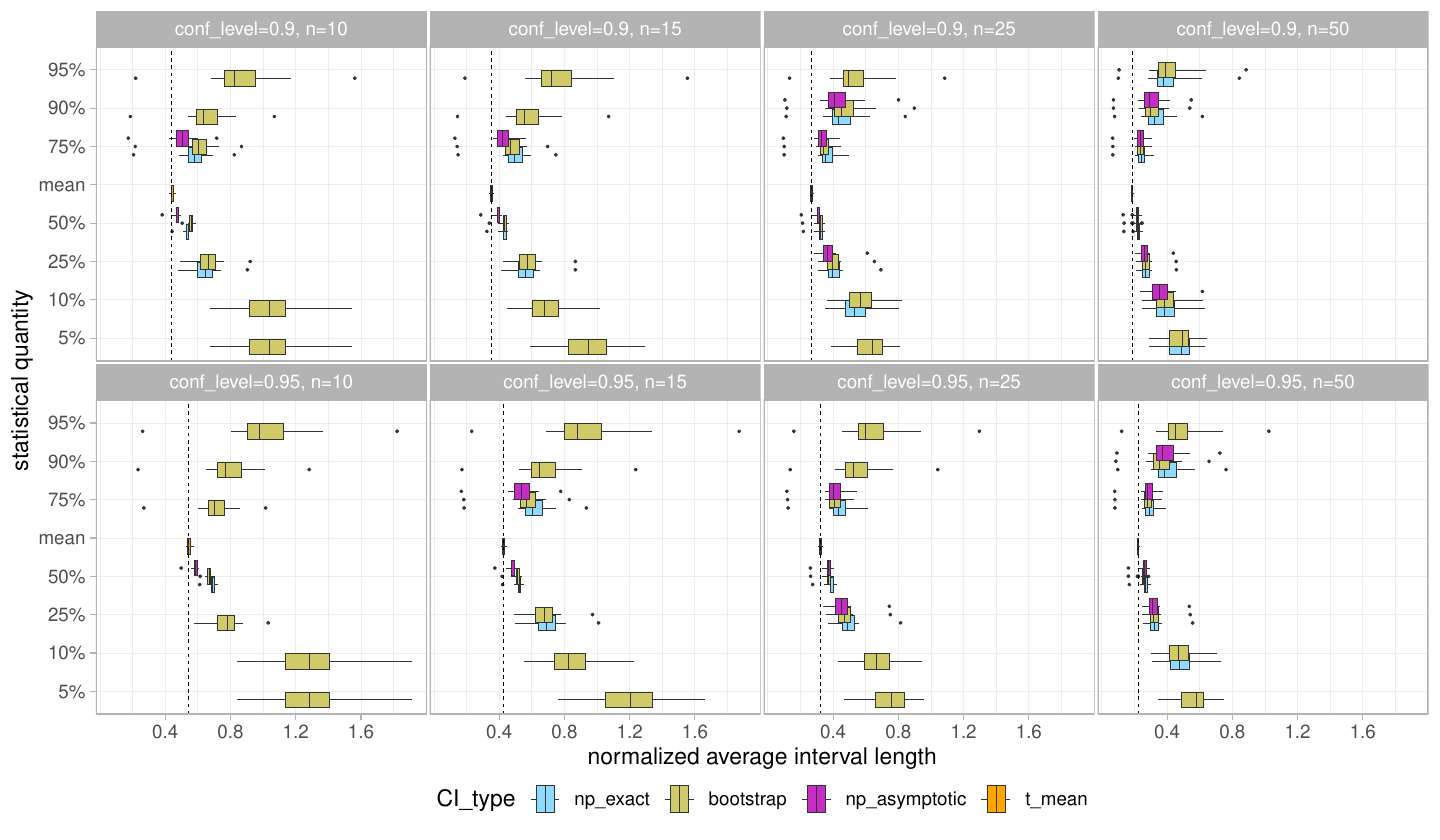}
    \caption{Normalized average interval length (reference: interdecile range) for different distributions for different types of estimated \acsp{ci} for quantile levels $5\%, 10\%, 25\%, 50\%, 75\%, 90\%, 95\%$ and the mean for confidence levels $1-\alpha = 0.90, 0.95$ and sample sizes $n=10, 15, 25, 50$. Shown results are based on $\num{2000}$ samples of size $n$ and $\num{2000}$ bootstrap samples. A total of 42 experiments is considered.}
    \label{fig:real-data-CI-avg-length}
\end{figure}

The results of different types of estimated \acsp{ci} for all experiments are shown in Figures~\ref{fig:real-data-CI-emp-coverage} and~\ref{fig:real-data-CI-avg-length}.
As there is no systematic and remarkable difference between the realized values for evaluation criteria of the estimated \acsp{ci} for the accuracy rate in the classification tasks (Simpsons characters, CIFAR10) and the RMSE in the regression task (superconductors), all use cases were aggregated into one plot.
That means, every single boxplot in Figures~\ref{fig:real-data-CI-emp-coverage}, \ref{fig:real-data-CI-avg-length} comprises a total of 42 experiments.

Figure~\ref{fig:real-data-CI-emp-coverage} shows the overall comparison of the empirical confidence level with the theoretical confidence level, represented by the red dashed line in each subplot. 
The empirical confidence level illustrates the frequentist interpretation of a \ac{ci}, meaning that the proportion of estimated intervals covering the true value of the quantity of interest should be close to the theoretical confidence level. 
This interpretation reflects the overall success rate, yet it remains impossible to identify which specific intervals include the true value.
A certain type of \ac{ci} (represented by color) performs well, if the corresponding boxplot is mainly located near the red dashed line. 
This is most easily seen in the subplots on the far right, which represent estimations for $n = 50$. 
For this sample size, the empirical confidence levels of the estimated \acsp{ci} closely align with the theoretical confidence levels.
In contrast, in the subplots on the far left for $n=10$, the difference appears mainly along the vertical direction. 
The effect, where quantile estimation is easier in the middle of a distribution than in the tails, is evident. 
Empirical confidence levels of \acsp{ci} for quantile levels around 50\% (e.g. 25\%, 75\%), are the only ones that align with the theoretical confidence level. 
With increasing sample size, \acsp{ci} for quantiles near the tails improve and become more closely aligned with the confidence level. 
It should also be noted that not all types of \acsp{ci} are valid for every combination of quantile level, confidence level, and sample size.
From the perspective of minimum sample size requirements, this issue was discussed in Section~\ref{sec:statistical-background}, see Tables~\ref{table:min-sample-size-exact-CI} and~\ref{table:min-sample-size-np-asymptotic-CI}. 
Although bootstrap \acsp{ci} lack such restrictions, they tend to perform poorly when other \ac{ci} types are invalid for the same sample size. 
However, despite a drop in the empirical confidence level, down to around 0.85, bootstrap \acsp{ci} could still serve as a practical alternative in cases where no other method is available. 

Another important aspect is the length of the estimated intervals. 
Generally speaking, a shorter \ac{ci} is preferred as long as the desired confidence level is achieved. 
Figure~\ref{fig:real-data-CI-avg-length} shows the average length normalized to the interdecile range (i.e.~the difference between 90\% quantile and 10\% quantile) of the corresponding distribution for all intervals based on $\num{2000}$ simulation runs.
For the interval length, there is no clear reference to compare with, as in the case of empirical coverage with the confidence level.
As the mean is very common to be used in practical applications, it seems reasonable to use the corresponding $t$-interval as reference. 
For comparison, Figure~\ref{fig:real-data-CI-avg-length} indicates the median level of the normalized average length of the \ac{ci} for the mean by the black dashed line. 
Due to the stability of the mean's estimate, there is minimal deviation in these \acsp{ci}' average lengths. 
This is not surprising, as the mean is well-known being a stable estimator under quite general conditions. 
Given the mean's stability, it is unlikely that a quantile \ac{ci} will have shorter length than the $t$-intervals for the mean with the same confidence level.
Therefore, quantile \ac{ci} average lengths that are comparable to those of the mean's \ac{ci} (black dashed line) are considered reasonable.
Figure~\ref{fig:real-data-CI-avg-length} shows that the average length of \acsp{ci} increases with higher confidence levels and decreases with larger sample sizes, an effect aligning with theoretical expectations.
Overall, for sample sizes up to~$25$, it seems reasonable to choose a maximum confidence level of 0.9 because higher confidence levels result in larger interval lengths and lower precision. 
Intervals for a confidence level of~$0.99$ were also calculated, and they showed good empirical coverage aligning well with the theoretical confidence level. 
However, these intervals were much longer: up to twice the interdecile range. 
Such long intervals are not practical for most applications, so this confidence level is not considered further.
For sake of completeness, the simulations as described above were run for a theoretical confidence level of 0.8 as well. 
But there was no benefit: while empirical coverage showed even more variation, the precision gain by shorter interval estimates was negligible. 
As a practical consequence, a confidence level of around 0.9 seems reasonable overall for sample sizes up to 25.

It is important to emphasize that checking the minimum sample size requirements for nonparametric \acsp{ci} provides valuable information about the quality of the bootstrap intervals that can be expected. 
Upper quantiles (beyond 75\% level) can be estimated more reliably than lower quantiles (below 25\% level).
Note that this asymmetry is reflected in the minimum sample requirements in Table~\ref{table:min-sample-size-np-asymptotic-CI}.

A useful observation is that the \ac{ci} for the mean (orange) produces short intervals with little variation. 
This is not surprising because the sample mean as estimator tends to stabilize quickly under quite general conditions and the normality assumption for the $t$-interval may be violated here. 
However, the application of goodness-of-fit tests (Lilliefors and Shapiro-Wilk) to the samples from the different empirical distributions show that normality cannot be strongly rejected. 
The $p$-values are mainly in the range of 0.2–0.25, with many around 0.5 or higher.
This result is due to the small sample sizes ($n=10$ or $n=15$), which are compatible with many distributions, including the normal distribution. 
A single exception here are the samples in the CIFAR10 use case from the left-skewed distribution in Figure~\ref{fig:experiments-classification-cifar10} (yellow color).
In this experiment, SGD is used as optimizer in connection with hyperparameter optimization. 
The \pvals of these small samples that are generated from this distribution are mainly below 0.1. 
With growing sample size ($n=25, 50$) these \pvals tend to zero and the normality property would be rejected.
However, for larger samples ($n=25$ or $n=50$), the CLT ensures that the $t$-interval behaves like an approximate interval that no longer depends on strict normality but only requires a sufficiently large sample size.

The bottom line so far: For sample sizes up to 25, it is reasonable to choose a maximum confidence level of~$0.9$ due to the reduced precision associated with larger interval lengths at higher confidence levels. 
While semiparametric bootstrap intervals can offer useful insights even with small sample sizes, this comes at the cost of longer \ac{ci} estimates. 
Especially with small samples, the $t$-interval for the mean provides a reasonable way to describe some aspects of the unknown distribution while accounting for uncertainty already.

\subsection{Validation}\label{sec:experiments-simulation}

While the real-data use cases provide insight into the practical behavior of the considered estimators under realistic conditions, the underlying distributions remain unknown. 
Furthermore, the assessment of statistical properties such as coverage and interval length can only be approximated.
To complement the empirical analysis above with a validation of the observed results, this section considers controlled simulation settings with known distributions. 

Previous work on quantile \acsp{ci}, such as \cite{hutson2002semiparamQuantileEstBT}, primarily considers distributions with unbounded support (e.g., normal, logistic, Laplace, Cauchy) or one-sided bounded support (e.g., log-normal, exponential). 
However, performance metrics in \ac{ml}, such as accuracy or F1 score, are typically defined on bounded intervals and may exhibit skewness.

To account for these characteristics, a left-skewed Beta distribution is considered. 
In addition, the results of the goodness-of-fit tests reported in Section~\ref{sec:experiments-real-data} suggest that a normal distribution can serve as a reasonable approximation in many cases. 
Therefore, a normal distribution is included as a simple baseline.
Furthermore, a mixture of normal distributions is considered to reflect potential bimodality. 
Such patterns were observed in some experimental settings, for instance due to irregularities in the training data. 
Although these cases were not suitable for a more detailed empirical analysis, they indicate that multimodal performance distributions may occur in practice and should be taken into account.

The parametrization of these distributions is guided by the empirical distributions from the real-data use cases, in particular by their extreme cases with respect to skewness and variance. 
This ensures that the validation covers challenging, but still realistic scenarios for quantile estimation in \ac{ml} evaluation.

\begin{figure}[ht!]
    \centering
    \includegraphics[width=0.85\linewidth]{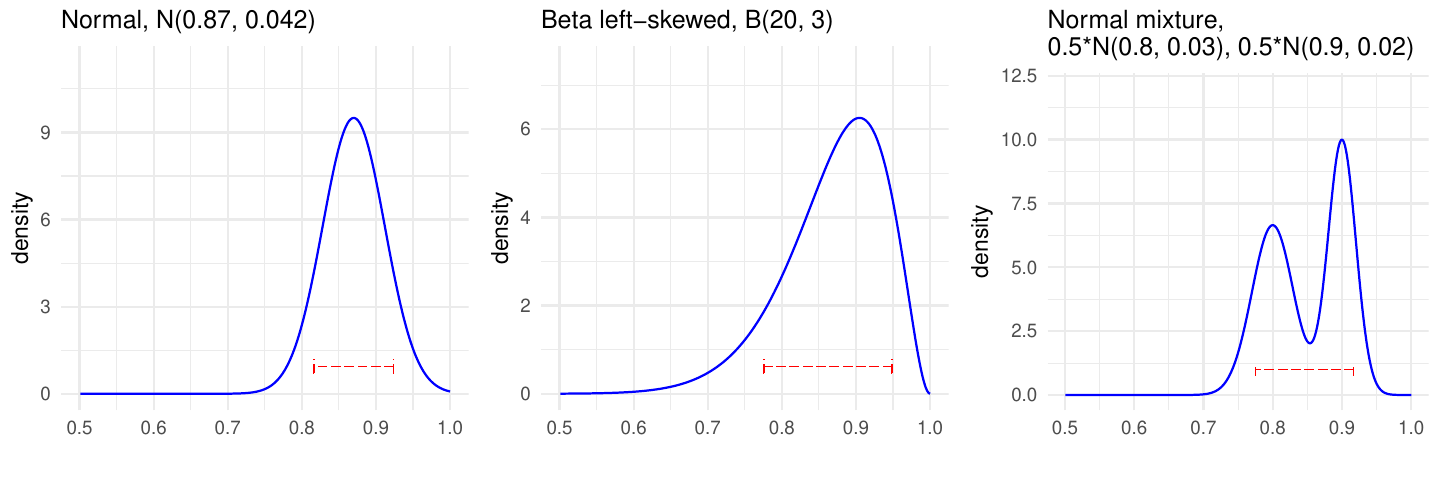}
    \caption{Densities of all considered distributions in simulation. Dashed horizontal red line: interdecile range (10\%, 90\%).}
    \label{fig:simulation-overview-densities}
\end{figure}

All these distributions including their parameters are shown in Figure~\ref{fig:simulation-overview-densities}.
Thereby, the normal distribution is parametrized such that most of its mass lies within the unit interval, representing a simple approximation frequently encountered in practice, despite the potential bounded nature of the underlying metrics.
For comparison reasons later on, the interdecile range from 10\% to 90\% quantile is marked by the red dashed line.

\subsubsection{Quantile Point Estimation}

Analogously to Section~\ref{sec:experiments-simulation}, the behavior of the point estimators introduced in Section~\ref{sec:Quantiles-and-point-estimation} is considered in terms of bias and RMSE using \num{2000} simulation runs. 
Within a single simulation run there are drawn random samples of size 10, 15, 25, 50 from each of the predefined distributions and the corresponding point estimates for the quantiles are computed.
In contrast to the real-data use cases, the true values of the estimated quantiles are known here. 

\begin{figure}[ht!]
    \centering
    \includegraphics[width=0.95\linewidth]{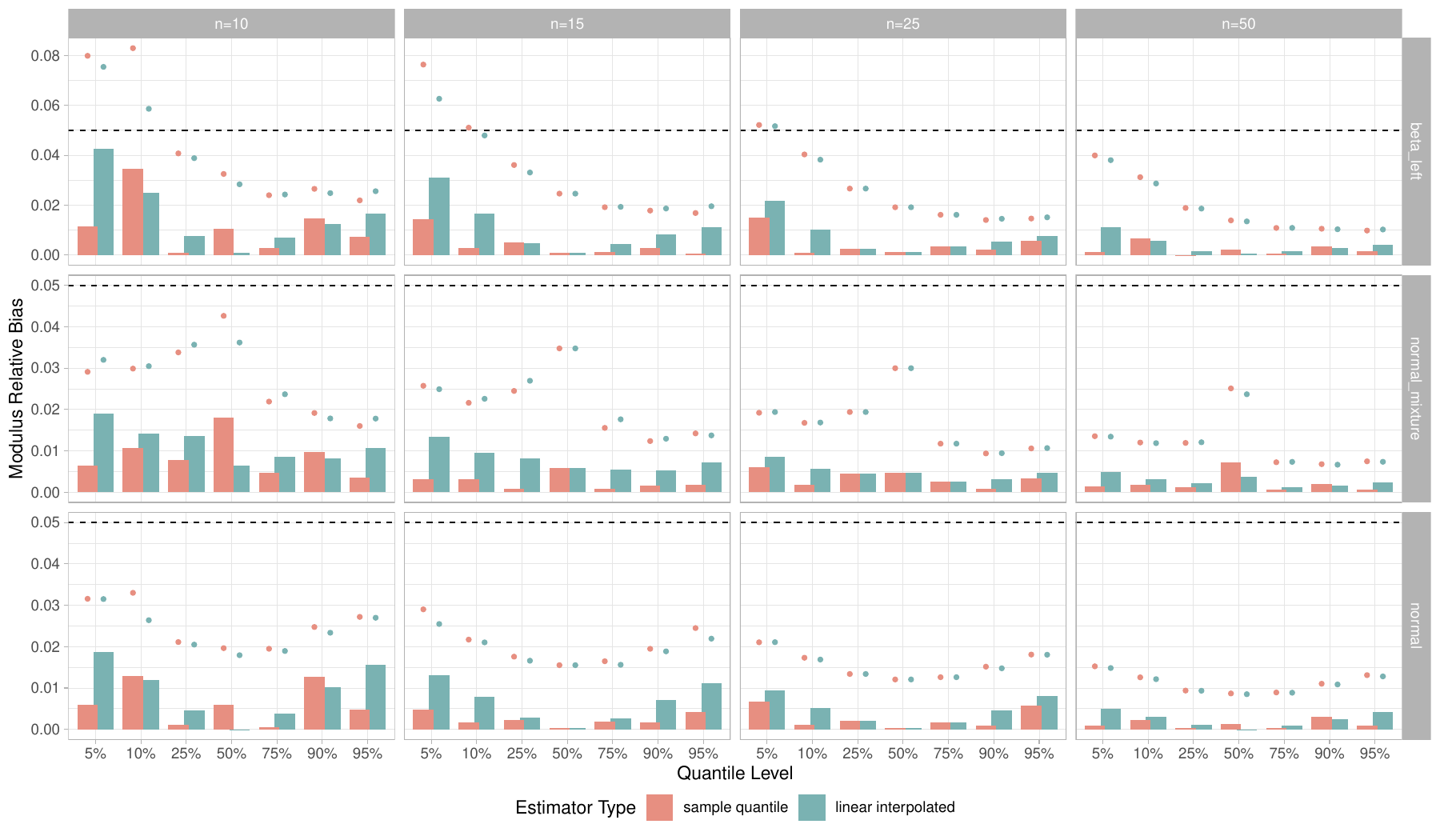}
    \caption{Average modulus relative bias of different quantile point estimators: sample quantile ($\hat{Q}$), interpolated quantile ($\hat{Q}_L$). Points are indicating the corresponding relative RMSE. The black dashed line marks the 5\% threshold. Shown results based on \num{2000} simulation runs for different quantile levels.}
    \label{fig:simulation-point-estimates-bias}
\end{figure}

Figure~\ref{fig:simulation-point-estimates-bias} illustrates the average of modulus bias and RMSE, normalized to the true quantile values, for various point estimators: sample quantile ($\hat{Q}$) and interpolated quantile ($\hat{Q}_L$), across different quantile levels. 
The modulus relative bias is displayed as a bar plot, with the corresponding relative RMSE represented by a dot. 
Colors are used to differentiate the estimators.
A horizontal black dashed line at the 5\% threshold is included for orientation.
Figure~\ref{fig:simulation-point-estimates-bias} shows the following key observations. 
As the sample size $n$ increases, both relative bias and RMSE decrease overall. 
Across the quantile levels, a (slightly skewed) U-shaped pattern is observed, with lower relative bias and RMSE for central quantiles and larger deviations in the tails. 
An exception is the normal mixture distribution, where mass concentrations away from the center make the estimation of central quantiles more challenging.

Compared to the real-data use cases (cf.\ Figure~\ref{fig:all-use-cases-point-estimates-bias}), the overall level of relative bias and RMSE is higher. 
This highlights the influence of the underlying distribution shape on quantile estimation. 
In particular, both the magnitude of bias and RMSE differ across distributions. 
Since the validation settings include more challenging distributional characteristics, the results can be interpreted as a conservative assessment of estimator performance.

Considering point estimates alone reveals the limitations of this perspective, as they do not provide information about the variability of the single estimates. 
This becomes particularly relevant for the bimodal distribution, where such irregularities are not apparent from point estimates alone. 
In contrast, the inspection of \acsp{ci} provides additional insights, for instance by revealing increased uncertainty or instability in the estimation.

\subsubsection{Quantile Interval Estimation}

This section applies the \acsp{ci} from Section~\ref{sec:non-parametric-CI} to the validation distributions from Figure~\ref{fig:simulation-overview-densities}.

The structural presentation is quite similar to Section~\ref{sec:experiments-real-data-interval-estimation}.
Again, all results presented here are based on $\num{2000}$ simulation runs and $\num{2000}$ bootstrap samples.
The only difference is, that the single distributions are considered separately.

\begin{figure}[ht!]
    \centering
    \includegraphics[width=0.98\linewidth]{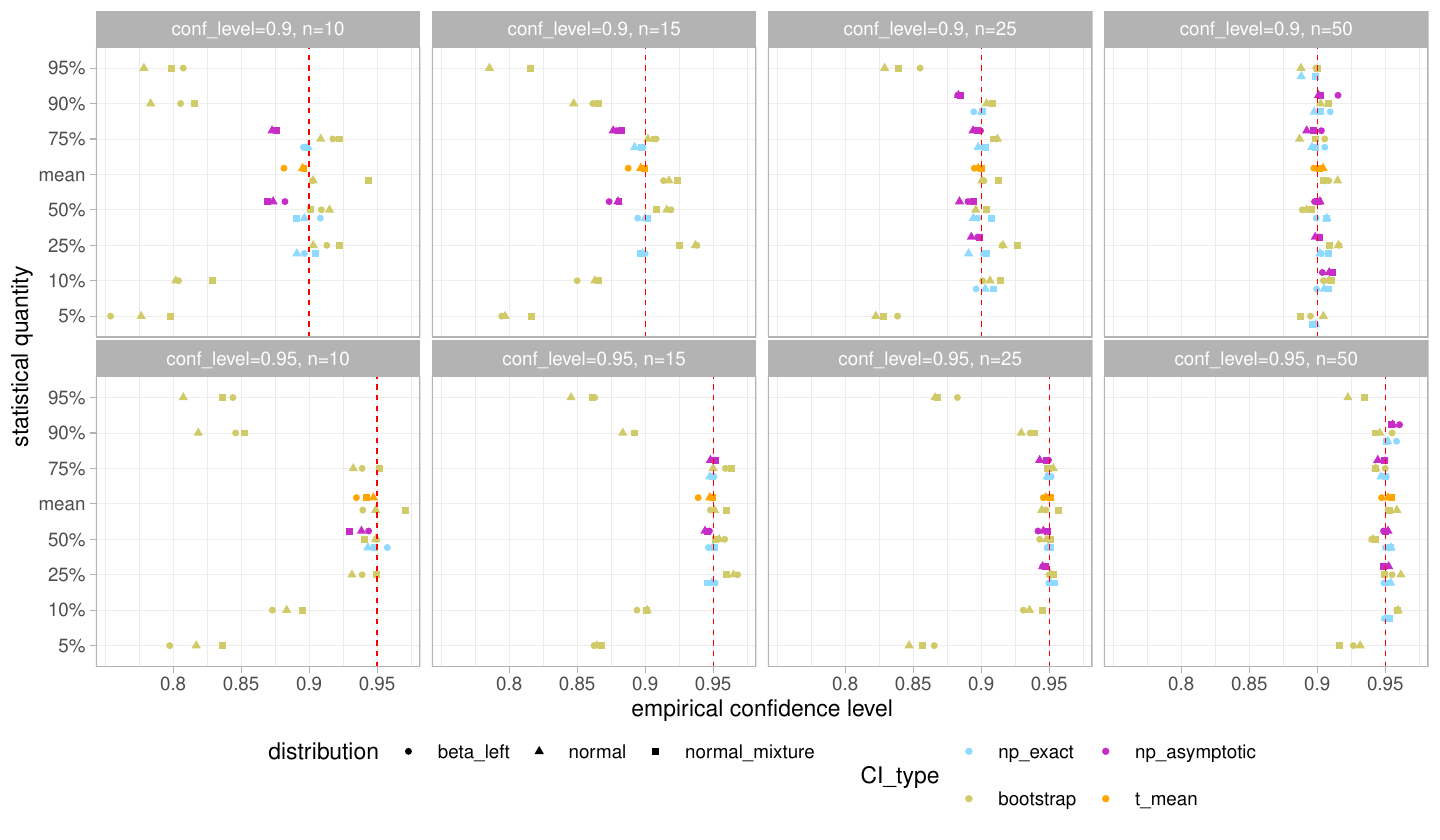}
    \caption{Empirical confidence level (along the horizontal axis) for different types of simulated \acsp{ci} for quantile levels $5\%, 10\%, 25\%, 50\%, 75\%, 90\%, 95\%$ and the mean for confidence levels $1-\alpha = 0.90, 0.95$ and sample sizes $n=10, 15, 25, 50$. Shown results based on \num{2000} simulation runs and \num{2000} bootstrap samples.}
    \label{fig:simulation-emp-coverage}
\end{figure}

\begin{figure}[ht!]
    \centering
    \includegraphics[width=0.98\linewidth]{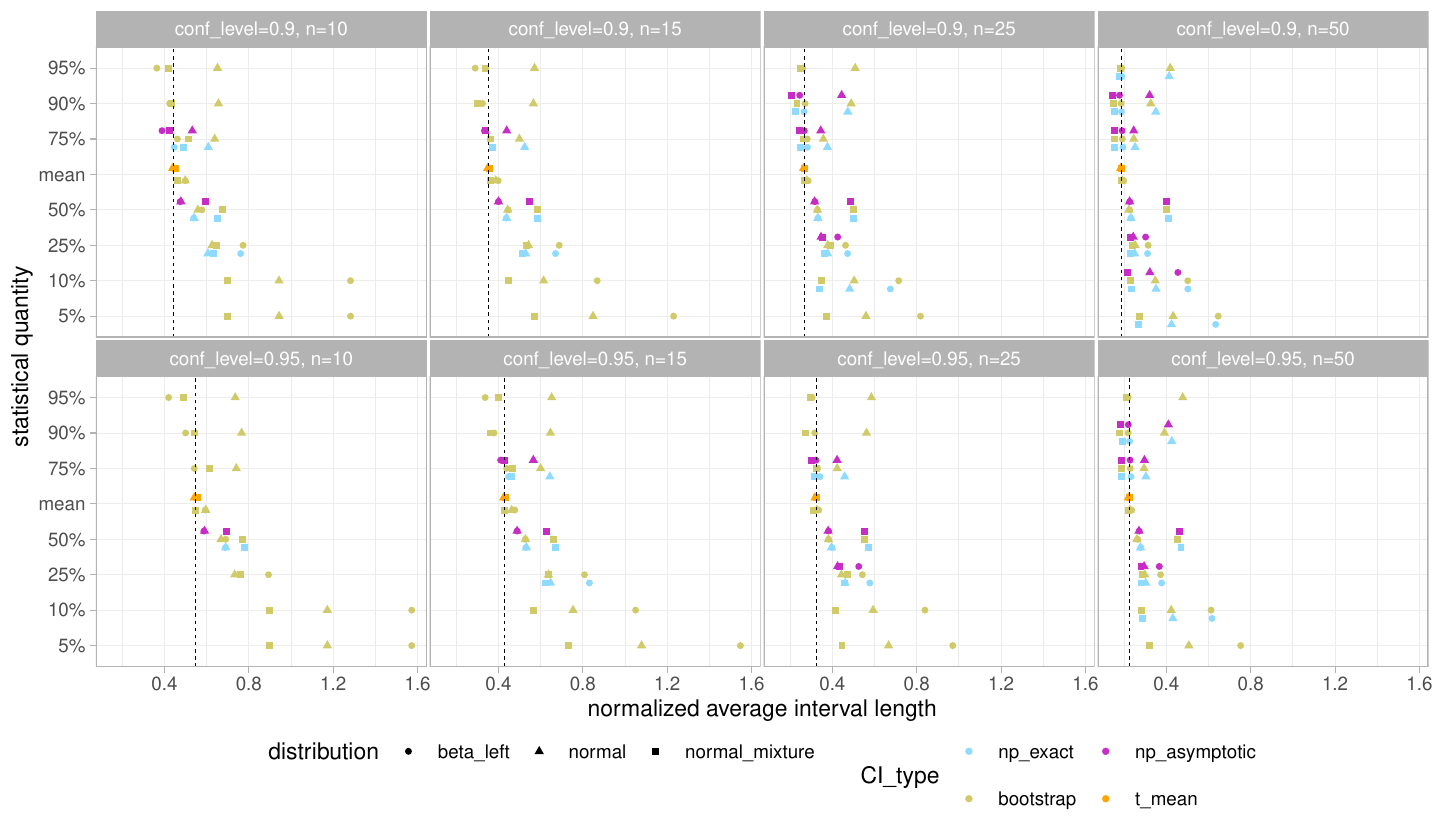}
    \caption{Normalized average interval length (reference: interdecile range) for different distributions for different types of simulated \acsp{ci} for quantile levels $5\%, 10\%, 25\%, 50\%, 75\%, 90\%, 95\%$ and the mean for confidence levels $1-\alpha = 0.90, 0.95$ and sample sizes $n=10, 15, 25, 50$. Shown results based on $\num{2000}$ simulation runs and $\num{2000}$ bootstrap samples.}
    \label{fig:simulation-avg-length}
\end{figure}

Figure~\ref{fig:simulation-emp-coverage} displays the empirical confidence levels of simulated \acsp{ci} for all the distributions from Figure~\ref{fig:simulation-overview-densities}. 
A single point in Figure~\ref{fig:simulation-emp-coverage} (and also Figure~\ref{fig:simulation-avg-length}) is characterized by:
\begin{enumerate}
    \item distribution, represented by plotting symbol%: circle, square, triangle
    \item \ac{ci} type, represented by color%: non-parametric exact, non-parametric asymptotic, bootstrap, $t$-interval for mean
\end{enumerate}

In Figure~\ref{fig:simulation-emp-coverage} the overall comparison with the theoretical confidence level, represented by the red dashed line in each subplot, is the key point of interest. 
If a certain type of \ac{ci} (represented by color) performs well, the points are located near the red dashed line. 
Roughly speaking, the overall impression of Figure~\ref{fig:simulation-emp-coverage} is very similar to Figure~\ref{fig:real-data-CI-emp-coverage} with the real-data use cases.
For practical reasons, differences among the considered distributions are more or less negligible in terms of empirical coverage probability.

More interesting here, is the average length of the resulting interval estimates.
Figure~\ref{fig:simulation-avg-length} shows the normalized average interval length for all distributions. 
Again, the normalization refers to the interdecile ranges, that are illustrated in Figure~\ref{fig:simulation-overview-densities}.
For comparison reasons, the length of the $t$-interval for the mean is marked by the black dashed line.
As a reminder, a shorter \ac{ci} is preferred as long as the desired confidence level is achieved. 
Figure~\ref{fig:simulation-avg-length} shows that the average length of \acsp{ci} increases with higher confidence levels and decreases with larger sample sizes, an effect aligning with theoretical expectations.

Further insights arise when considering different quantile levels. 
In skewed distributions (cp.\ beta\_left), quantiles located in the long tail are estimated with higher uncertainty, resulting in longer \acsp{ci}. 
For the left-skewed Beta distribution, this primarily affects the lower quantiles. 
Thus, the generally increased estimation uncertainty in the tails is further amplified by skewness, leading to longer intervals.
However, left-skewness is more difficult to detect based on interval lengths alone, since extreme quantiles are inherently harder to estimate even in symmetric settings. 
Conversely, right-skewed distributions tend to exhibit larger intervals for higher quantile levels. 
From a risk-oriented perspective, this is particularly relevant when extreme quantiles coincide with the long tail of a skewed distribution, as both effects jointly increase estimation uncertainty.

Another interesting aspect when considering different quantile levels is bimodality.
Bimodality may be indicated by relatively larger intervals for the median compared to both lower and upper quantiles, as e.g.\ the 25\% and 75\% quantiles.
This effect is consistent in Figure~\ref{fig:simulation-avg-length} for the normal mixture (square symbol) for a sample size of 15 and above.

Regarding the different estimators, the asymptotic nonparametric estimator performs surprisingly well for small sample sizes. 
But it comes with the drawback that the minimum sample size to estimate a \ac{ci} for the 25\% quantile is slightly higher than the other estimators.
This drawback can be mitigated by using the sign transformation as mentioned earlier.
Overall, there is no clear outperformer among the estimators and the interval length is mainly driven be quantile level and distribution shape.

Another useful observation is that the \ac{ci} for the mean (orange) produces short intervals with little variation. 
This is not surprising because the sample mean as estimator tends to stabilize quickly under quite general conditions.
Although, the normality assumption for the $t$-interval is violated here, as most distributions are not normal. 
However, the application of goodness-of-fit tests (Lilliefors and Shapiro-Wilk) to the samples from different distributions show that normality cannot be strongly rejected. 
The $p$-values are mainly in the range of 0.2–0.8, with many around 0.5 or higher.
This result is due to the small sample sizes ($n=10$ or $n=15$), which are compatible with many distributions, including the normal distribution. 
For larger samples ($n=25$ or $n=50$), the CLT ensures that the $t$-interval behaves like an approximate interval that no longer depends on strict normality but only requires a sufficiently large sample size.
As a result, especially with small samples, the \ac{ci} for the mean provides a reasonable way to account for uncertainty already.

Finally, some comparisons of the simulation results can be made with other existing research contributions.
A first comparison is done with the simulation results based on generalized bootstrap from \cite{wang2010comparisonBTgeneralizedBTQuantileEst}.
Although slightly different theoretical distributions are considered therein, they remain comparable (e.g., Beta distributions with varying parameters).
For small sample sizes ($n=10, 15$), the empirical coverage probabilities of the generalized bootstrap are slightly lower than those of the semiparametric bootstrap but with shorter average interval lengths. 
Furthermore, \cite{wang2010comparisonBTgeneralizedBTQuantileEst} also explores parametric bootstrap methods, which could serve as a best-case bound for bootstrap \ac{ci} estimation. 
Another comparison can be made with the simulation results from \cite{nagaraja2020distributionFreeApprox}, whose detailed numbers are provided in a supplementary document available on the publisher's website. 
The study by \cite{nagaraja2020distributionFreeApprox} investigates many various approaches for constructing approximate \acsp{ci} for quantiles, whereby considering sample sizes of $n=10, 50, 100, 1000$. 
Although the paper primarily focuses on theoretical distributions that are less relevant for \acsp{tmoi} with bounded support in \ac{ml}, such as the normal, gamma, Pareto, and $t$-distributions, it allows for comparison of results for the normal distribution. 
For the sample sizes of $n = 10$ and $n = 50$, the empirical coverage and average lengths of the estimated \acsp{ci} show a comparable order, particularly for the reasonable results.
Overall, these comparisons suggest that the approaches presented here give acceptable \ac{ci} results, at least leaving limited potential for substantial improvement by using other approaches.
\section{Conclusion and Further Work}
\label{sec:conclusion}

This work investigates the empirical distribution of machine learning performance metrics by applying quantile-based point and interval estimation techniques. 
In contrast to standard evaluation practices that rely on single aggregated values, performance is explicitly modeled as a random quantity induced by stochastic elements of the training process.
Such stochasticity aligns with experimental practice in the natural sciences, where repeated measurements under controlled conditions provide reliable insights.
In the present context, repeated training runs enable statistical inference on model performance and allow for a more detailed analysis of variability.
Moreover, the use of confidence intervals and the explicit consideration of estimation error enable systematic sample size planning to meet predefined accuracy requirements, which is standard practice in the natural sciences.

An important benefit of using quantiles is the formulation and verification of performance requirements in probabilistic terms. 
For instance, statements such as `the accuracy drops below 90\% with probability at most 10\%' or `the RMSE exceeds a threshold $y$ with probability at most 5\%' can be assessed directly based on quantile estimates and corresponding confidence intervals. 
This provides a natural link to a risk-oriented interpretation of model performance.

None of the considered approaches requires parametric assumptions about the distribution of the \ac{tmoi}, making them broadly applicable across model classes and performance metrics. 
For reference, the classical $t$-interval for the mean is included as a baseline.
Real-data experiments for classification and regression tasks, together with validation studies, show that the shape of the underlying distribution has a strong impact on estimation quality. 
In particular, deviations from symmetry, such as skewness or multimodality, affect the reliability of quantile estimates. 
As a consequence, quantile estimates in combination with interval lengths provide indirect information about such distributional characteristics.

Regarding sample size, values in the range of $n = 15$ to $25$ are sufficient to obtain informative confidence intervals for central and moderately extreme quantiles (up to approximately the 90\% and 10\% levels). 
Across all considered sample sizes and confidence levels, the nonparametric asymptotic confidence interval provides a favorable trade-off between applicability, statistical quality, and simplicity, making it a practical default choice in many settings. 
A sample size of $n = 10$ can still provide useful insight for central quantiles (e.g.\ the first and third quartiles), although with reduced statistical reliability. 
In such small-sample settings, semiparametric bootstrap intervals offer an alternative when other nonparametric methods are not applicable due to sample size constraints. 
Although the resulting coverage is lower and intervals are wider, they still provide useful indications of uncertainty.

\bigskip
The results for small sample sizes ($n \leq 10$) indicate clear limitations, particularly for the estimation of extreme quantiles, and suggest several directions for future research.
A first direction is the improvement of semiparametric bootstrap approaches. 
In particular, extensions based on tail extrapolation and smoothed bootstrap methods, as proposed in \cite{wei2015quantileEstimatVSCIcoverage}, show promising results in terms of coverage accuracy and interval length, even for extreme quantiles under small sample sizes. 
However, these methods are technically more demanding and have not yet been systematically evaluated for distributions with bounded support, which are common in machine learning applications.
A second direction is the use of importance sampling techniques for improving the estimation of confidence interval bounds, as discussed in \cite{hu2008bootstrapQuantileEstImportanceSampling}. 
While existing work focuses on other statistical quantities, the transfer of these ideas to quantile estimation in the present context remains an open question.
Finally, the distinction between single and simultaneous confidence intervals for multiple quantiles represents a relevant extension. 
Since quantiles are inherently dependent, simultaneous inference procedures, as discussed in \cite{hayter2014simultaneousQuantileCI}, may provide a more consistent framework when multiple quantile levels are of interest.

\bigskip
For practical applications, the uncertainty perspective may reveal different aspects, especially when choosing among a few competing models at the end of a development process.
The width and position of confidence intervals serve as direct indicators of uncertainty and can support decision-making in applied machine learning settings.
Overall, the results demonstrate that distributional evaluation of performance metrics is feasible under realistic constraints and provides relevant additional information beyond standard mean-based reporting.

\bigskip
All relevant code and implementations of the non-standard \ac{ci} estimators will soon be made available on \url{https://github.com}.

%hints that assumption of normal distribution seems reasonable in some cases, but there is hardly any possibility to substantially justify this assumption. in contrast, the presented approaches here only assume some continuous distribution. consequently, an assumption weak or free approach can be interpreted as a conservative perspective, whereas the normal distribution can be seen as a more liberal perspective. 

% potential extensions:
% - consider point estimates of quantiles (not possible for nonparametric exact intervals)
%     - question: how to define point estimation for the semiparametric bootstrap? Maybe, median of bootstrap distribution or mean (already available)
% - analyze bias and variance of those based on the existing simulations (all necessary data is available)
% - discussion of undershot confidence level already contained implicitly: choosing conf level of 0.8 leads only to slightly smaller intervals, thus undershooting by only 1 or 2 percentage points probably does not have practical relevant implications on length of resulting intervals. Thus, there is hardly any waste of confidence level.

\section*{Acknowledgments}
\label{sec:acknowledgments}
The authors acknowledge the financial support by the Federal Ministry of Education and Research of Germany and by Sächsisches Staatsministerium für Wissenschaft, Kultur und Tourismus in the programme Center of Excellence for AI-research `Center for Scalable Data Analytics and Artificial Intelligence Dresden/Leipzig', project identification number: ScaDS.AI

The authors gratefully acknowledge the computing time made available to them on the high-performance computer at the NHR Center of TU Dresden. 
This center is jointly supported by the Federal Ministry of Education and Research and the state governments participating in the NHR (www.nhr-verein.de/unsere-partner).

\printbibliography

\end{document}